\documentclass{article}

\usepackage{PRIMEarxiv}

\usepackage[utf8]{inputenc} % allow utf-8 input
\usepackage[T1]{fontenc}    % use 8-bit T1 fonts
\usepackage{hyperref}       % hyperlinks
\usepackage{url}            % simple URL typesetting
\usepackage{booktabs}       % professional-quality tables
\usepackage{amsfonts}       % blackboard math symbols
\usepackage{nicefrac}       % compact symbols for 1/2, etc.
\usepackage{microtype}      % microtypography
\usepackage{lipsum}
\usepackage{fancyhdr}       % header
\usepackage{graphicx}       % graphics
\graphicspath{{media/}}     % organize your images and other figures under media/ folder
\usepackage{amsmath}
\usepackage{gensymb}
\usepackage{longtable}
\usepackage{subcaption}
\usepackage{pdfpages}

\title{Heatwave increases nighttime light intensity in hyperdense cities of the Global South: A double machine learning study
}

\author{
  Ramit Debnath \\
  University of Cambridge \\
  Cambridge, UK\\
  \texttt{\{rd545\}@cam.ac.uk} \\
   \And
  Taran Chandel \\
  University of Cambridge \\
  Cambridge, UK\\
  And 
  University of Edinburgh \\
  Edinburgh, UK\\
  \AND
  Fengyuan Han \\
  University of Cambridge \\
  Cambridge, UK\\
 \And
  Ronita Bardhan \\
  University of Cambridge \\
  Cambridge, UK\\
}

\begin{document}
\maketitle

\begin{abstract}
Heatwaves, intensified by climate change and rapid urbanisation, pose significant threats to urban systems, particularly in the Global South, where adaptive capacity is constrained. This study investigates the relationship between heatwaves and nighttime light (NTL) radiance, a proxy of nighttime economic activity, in four hyperdense cities: Delhi, Guangzhou, Cairo, and São Paulo. We hypothesised that heatwaves increase nighttime activity. Using a double machine learning (DML) framework, we analysed data from 2013 to 2019 to quantify the impact of heatwaves on NTL while controlling for local climatic confounders. Results revealed a statistically significant increase in NTL intensity during heatwaves, with Cairo, Delhi, and Guangzhou showing elevated NTL on the third day, while São Paulo exhibits a delayed response on the fourth day. Sensitivity analyses confirmed the robustness of these findings, indicating that prolonged heat stress prompts urban populations to shift activities to night. Heterogeneous responses across cities highlight the possible influence of urban morphology and adaptive capacity to heatwave impacts. Our findings provide a foundation for policymakers to develop data-driven heat adaptation strategies, ensuring that cities remain liveable and economically resilient in an increasingly warming world.
\end{abstract}

% keywords can be removed
\keywords{Heatwaves \and double machine learning \and hyperdensity \and Global South \and nightlight \and cities}

\section{Introduction}
Heatwaves are becoming increasingly frequent and severe, posing significant threats to urban quality of life. This trend is projected to worsen due to climate change and rapid urban expansion, particularly in the Global South, where adaptive capacity is often limited. Understanding and mitigating the impacts of heatwaves in these regions is critical to reducing population vulnerability and improving resilience. Urban heatwaves are driven by a combination of factors, including the urban heat island (UHI) effect, which arises from human activities (e.g., transportation, air conditioning, and building operations), urban infrastructure (e.g., asphalt pavements, narrow streets, and high-rise buildings), and broader climatic changes such as rising global temperatures and changing atmospheric circulation patterns \cite{cheval2024systematic, li2024cooling}.

At a macro-level, heatwaves have been shown to significantly impair economic and labour productivity. For example, daytime work efficiency can decrease by up to 10\% during extreme heat events \cite{kjellstrom2016heat}, with particularly pronounced impacts in warmer climates. Rapid and often unplanned urbanisation, a characteristic of many cities in the Global South, exacerbates heat-related discomfort, potentially triggering urban migration and increasing public health risks, including mortality \cite{yuan2024unraveling, li2024heatwave, debnath2024better}. Although cities in low-income regions face climate stresses similar to those in wealthier areas, they often lack the resources and infrastructure to mitigate these challenges effectively \cite{yuan2024unraveling, choudhary2019can, debnath2023lethal}. Consequently, increasing the vulnerability of the population to heatwave events. 

Existing research, primarily focused on Europe, North America and parts of Asia, has demonstrated that heatwaves intensify the effect of UHI, particularly at night, when the temperature difference between urban and suburban areas becomes more pronounced \cite{an2020observational, burger2021modelling, li2024cooling}. Nighttime heatwaves also pose increased health risks, as reduced adaptive capacity during sleep can lead to increased mortality from prolonged exposure to elevated temperatures \cite{laaidi2012impact}. In addition, there has been growing literature on exploring the economic impacts of heatwaves on an aggregated scale. For example, Babii et al. (2023) have integrated ML into econometric models to improve the accuracy of economic forecasts, particularly in the context of climate-induced fluctuations \cite{babii202410}. Similarly, Buster et al. (2024) developed computationally efficient ML methods to improve urban temperature estimates, which are critical to evaluating the economic benefits of heat mitigation strategies \cite{buster2024tackling}. Chakraborty and Stokes (2023) introduced an ML framework that adapts to city-specific nighttime light signatures (NTL), allowing the tracking of urban changes and providing insight into economic dynamics \cite{chakraborty2023adaptive}.  However, there is limited evidence that economic activities are shifting toward night during heat wave events in urban areas (see Section 2 for a detailed review of the literature). 

The primary objective of this study is to establish a robust association between heatwaves and nighttime light (NTL) radiance in hyperdense urban areas of the Global South. To achieve this, we employ a double machine learning (DML) approach, which combines the flexibility of nonparametric ML models with robust causal inference techniques. DML, introduced by Chernozhukov et al. (2018), enables the estimation of causal effects while controlling for confounders and providing valid confidence intervals \cite{chernozhukov2018double, fuhr2024estimating}. By applying DML, our aim is to establish a direct link between heatwave events and variation of NTL intensities while controlling for local climatic conditions to enable better heat-resilience planning in conditions where climate adaptive capacities are constrained. 

Our analysis focusses on four hyperdense cities in the Global South: Delhi (India), Guangzhou (China), Cairo (Egypt), and São Paulo (Brazil). These cities were selected for their heterogeneous climate conditions, sociodemographic characteristics, and rapid urbanisation, making them ideal case studies to understand the diverse impacts of heatwaves on urban systems. NTL serves as a proxy for human activity and energy use, making it a valuable indicator of urban productivity and adaptation during extreme heat events \cite{chakraborty2023adaptive}. By examining these cities, we aim to provide insights into how heatwaves influence NTL in different urban contexts when confounded with heterogenous climatic factors (see Table 1). The relationship between heatwaves and NTL established in this study has significant implications for planning for heat wave adaptation in areas of limited resources. Specifically, our findings establish that heatwaves influence NTL intensity values, indicating a change in urban nighttime activities. This relationship can help policymakers and urban planners identify areas where heatwaves disproportionately disrupt urban activities, including economic productivity. This information can guide targeted heatwave interventions and implement heat-resilient urban design. Furthermore, by understanding how heatwaves alter NTL, cities can better allocate limited resources to mitigate the impacts of extreme heat, thus reducing vulnerability and enhancing resilience in the face of a warming climate. Ultimately, this study contributes to a broader understanding of how heat waves affect urban systems in the Global South, providing an evidence-based foundation for adaptive strategies that balance economic growth with climate resilience.

The remainder of this paper is structured as follows. Section 2 provides a detailed background on the relationship between NTL and urban productivity, current approaches to measuring productivity shifts during heatwaves, and the limitations of NTL data. Section 3 describes the data acquisition, preprocessing and modelling approaches, including the DML framework, the setup of the treatment effect model, and the robustness checks using sensitivity analysis. Section 4 presents the results, and Section 5 discusses their implications for heat-resilient urban planning, climate adaptation, and policy in the Global South.

\section{Background}

\subsection{Nighttime light (NTL) and urban productivity}
Nighttime light (NTL) data have emerged as a powerful tool in socio-spatial research, offering unique insights into urban growth, economic activity, and infrastructure dynamics. Due to its direct link to human activities, NTL is widely used to characterise urban expansion and development \cite{zheng2023nighttime, levin2020remote}. Unlike daytime satellite imagery (e.g. Landsat or SPOT), NTL data provide a greater contrast between urban and rural areas, making it particularly effective for delineating the boundaries of built-up areas \cite{liang2017feasibility}. This characteristic has established NTL as one of the main data sets for mapping urban expansion. For example, as early as 1979, Croft demonstrated the utility of DMSP-OLS NTL images to map major cities in the eastern United States, highlighting their potential to represent economic growth and prosperity \cite{croft1979brightness}. Since then, NTL data have become a cornerstone in urban research, offering a unique lens through which to study socioeconomic and environmental phenomena.

Nighttime light (NTL) data possesses the capability to encompass a diverse array of socioeconomic and environmental parameters. This richness of information proves instrumental in informing and guiding decision-making processes. Traditional metrics such as Gross Domestic Product (GDP), population and energy consumption are often recorded at administrative levels, which can be coarse in spatial dimensions, difficult to obtain, and unsuitable for granular analysis \cite{bennett2017advances, mccord2022nightlights, zhou2022city}. In contrast, NTL data provide timely, spatially explicit, and consistent measurements of human activity, overcoming the delays and limitations associated with conventional data collection methods, which often suffer from lags of 1-2 years \cite{zhang2015estimating}. Henderson et al. \cite{henderson2003validation} and Pinkovskiy and Sala-I-Martin \cite{pinkovskiy2016lights} have shown that NTL effectively represents urban economic activity, facilitating its use in socioeconomic studies.

The application of NTL data in urban studies can be broadly categorised into three streams. The first stream focusses on establishing correlations between the intensity of NTL and socioeconomic variables such as GDP \cite{chen2011using, wang2022global} and population \cite{stathakis2018seasonal}. These studies have consistently shown that the NTL radiance is a reliable indicator of economic activity, making it a valuable tool for monitoring urban productivity and development. The second stream of research emphasises the development of statistical and machine learning models to explore the relationship between NTL radiance and variables of interest. Commonly used models include linear regression, panel regression, geographically weighted regression, exponential growth models, and random forest \cite{bennett2017advances, doll2010estimating, imran2019spatial, wang2021modeling}. More recently, deep learning techniques, such as convolutional neural networks (CNNs), have been employed to predict socioeconomic indicators such as GDP on finer spatial scales using NTL data as input \cite{sun2020estimation}. These advances highlight the growing sophistication of NTL-based analyses and their potential to capture complex urban dynamics. The third stream of research addresses the issue of scale, which is critical for generalising findings across different spatial contexts. Efforts have been made to bridge the gap between the training scale of models and their implementation scale, either by scaling from plot-level data to pixel-level data or by scaling from national or provincial scales to city or pixel levels \cite{rybnikova2016estimating, zhang2011mapping}. These methodological innovations have significantly improved the applicability of NTL data in diverse urban planning and governance contexts.

Within the realm of heatwaves and urban productivity, we hypothesise that NTL data can offer a distinctive opportunity to evaluate the effects of intense heat events on economic activities and infrastructure reliability.

\subsection{Heatwave-induced productivity shifts and its measurement}

Several empirical studies revealed that heatwaves can affect the economy by lowering labour productivity and increasing mortality risks \cite{somanathan2021impact, schlenker2009nonlinear, miller2021heat}. Between 1980 and 2017, heatwaves have become prolonged, exposing urban populations to unusually high heat stress, which accounted for approximately five percent of the economic losses in Europe \cite{worldbank2020analysis}. A prominent contributor to the exacerbation of heatwaves in cities is the Urban Heat Island (UHI) effect, which is sensitive to urban land use and the characteristics of the built form \cite{cheval2024systematic}.

There is a growing literature on the attribution of heatwave events to economic and productivity losses. For example, Costa et al. \cite{costa2016working} developed a comprehensive cost methodology that integrates urban climate modelling with labour productivity and economic production, revealing significant variability in the economic impacts of heat. These impacts are influenced by production characteristics, such as the elasticity of substitution between capital and labour, as well as the relative size of different economic sectors. Their findings estimate that, in the absence of adaptation measures, total economic losses during a warm year in the far future (2081–2100) could range from 0. 4\% of the gross value added in cities such as London, UK to up to 9. 5\% in cities such as Bilbao, Spain. Miller et al. \cite{miller2021heat} evaluated a global temperature dataset from 1979 to 2016, concluding that the potential damage to the agricultural industry could be five to ten times greater than prior estimates, with manufacturing and construction also highly vulnerable to heat-related impacts.

In addition, an expanding body of literature explores regional differences in the economic impacts of urban heat waves, highlighting concerns about environmental injustice. Callahan et al.\cite{callahan2022globally}, quantified the effects of extreme urban heat and found that human-induced increases in heatwaves (through anthropocentric emissions) have disproportionately reduced economic output in poorer tropical regions. Similarly, Garcia-Leon and colleagues \cite{garcia2021current} highlighted significant variations in economic losses from heatwaves across different spatial units, emphasising the heterogeneity of regional impacts. These disparities are driven by the interplay of geographical and social factors, with urban green infrastructure emerging as a critical determinant. For example, a recent study showed that the uneven spatial distribution of urban green infrastructure can exacerbate economic inequality during heatwaves \cite{zhou2021urban, li2024cooling}. 

According to \cite{cheval2024systematic}, the majority of urban heatwave research is focused on Europe, America, and Asia, with almost 40\% of the studies focused on Europe and America alone. This regional bias has resulted in the under-representation of rapidly urbanising cities, particularly those in the Global South. Expanding the geographical scope of urban heatwave studies is essential to develop equitable and globally relevant strategies to mitigate the socioeconomic impacts of rising temperatures and to ensure that no region is left behind in climate adaptation efforts.

Recently, a variety of econometric and ML models have been used to explore the relationship between urban heat waves and economic losses. For example, \cite{babii2023econometrics} highlighted the integration of ML methods into econometric models to improve the accuracy of economic forecasts under climate projections. These advanced models enable the analysis of high-dimensional data, providing a more nuanced understanding of how urban heat stress affects economic productivity. From a dynamic modelling perspective, studies have also used time series models such as ARX and ARMAX to examine temporal changes and forecast urban heatwaves \cite{chen2022estimating, gustin2018forecasting}. However, a gap remains in the literature related to the assessment of the dynamic interactions between urban heatwaves and economic growth or activities. Bridging this gap is essential to designing better adaptive frameworks for reducing the economic consequences of heatwaves in cities, particularly in the Global South.

ML techniques have been increasingly applied to assess the vulnerability of urban heat and its economic implications. For example, \cite{buster2024tackling} developed computationally efficient ML methods that enhance urban temperature estimates, which are critical for evaluating the economic benefits of heat mitigation strategies. Similarly, \cite{redding2024quantitative} reviewed quantitative urban models that analyse the spatial organisation of economic activity within cities, offering valuable information on how urban heat islands influence economic outcomes. Chakraborty and Stokes \cite{chakraborty2023adaptive} introduced an ML framework that adapts to city-specific NTL signatures, allowing the tracking of urban changes and providing insight into economic dynamics. This approach employs neural networks to forecast NTL patterns, allowing for the detection of deviations that signal infrastructure or economic changes.

\subsection{Critique of nighttime light data}

Despite the widely recognised potential of NTL data for urban mapping and measuring economic activities, its accuracy has been challenged by the blooming effect, which causes urban areas to appear larger than they actually are \cite{small2005spatial}. Addressing this issue has remained a central focus in NTL-based urban mapping research. In the early stages, researchers used the proportion of detected NTL signals relative to total cloud-free observations as an indicator to mitigate the blooming effect. The studies aimed to establish an empirical threshold for this percentage to filter out extraneous signals while maintaining consistent urban boundaries \cite{imhoff1997technique}. The underlying assumption was that the "blooming pixels" were transient and thus had lower detection rates \cite{henderson2003validation}. However, the precision of urban mapping was highly sensitive to the chosen threshold, and even within the same study area, empirical percentage thresholds varied significantly between studies \cite{elvidge1997relation, imhoff1997technique}. This variability highlights the challenges in developing a universal threshold and underscores the need for more robust methods to improve the reliability of NTL-based urban mapping. 

Another limitation of NTL data lies in the generalisability and scales of the models studied. Model selection can be highly dependent on the variables analysed and the specific study areas, particularly when these areas differ in socioeconomic conditions, outdoor lighting regulations, and population density. This variability raises concerns about the generalisability of the models, as those that perform well on a global scale may produce significant errors when applied at regional levels \cite{ma2018multi}. Furthermore, the use of different scales in the studies further complicates the generalisability of the models researched. Many studies train models at one scale and then apply them to scenarios with different spatial scales, implicitly assuming that relationships modelled at one spatial scale (e.g., national) can be directly applied to another (e.g., pixel). However, this assumption may not hold in many contexts. For example, when estimating the population using NTL data, models are typically developed based on the correlation between total NTL intensity and total population at the national level \cite{zheng2023nighttime}. These models are then used to estimate the population at the pixel level on the basis of the NTL intensity of individual pixels. However, this approach overlooks a critical issue: most of the total intensity of NTL originates from brightly lit urban areas, while dimly lit or dark rural regions - although they host a significant proportion of the population - are often under-represented in the models \cite{doll2010estimating}. This discrepancy highlights the need for more nuanced approaches to ensure the accuracy and applicability of NTL-based models across diverse spatial scales and contexts.

Lastly, several studies have concluded that relying solely on NTL data is insufficient to accurately measure GDP at subnational levels. The limitation of using NTL as the only predictor arises from its inability to represent areas without visible nighttime lights. Specifically, not all economic activities, such as agriculture and forestry, exhibit increased NTL emissions as they expand \cite{keola2015monitoring}. To address this shortcoming, researchers suggest incorporating additional data sources to improve GDP estimation. For example, \cite{keola2015monitoring} developed a model that integrates land cover data with NTL to estimate agricultural and non-agricultural economic growth at national and subnational levels. Similarly, \cite{yue2014estimation} introduced a systematic approach for a more precise estimate of GDP by combining DMSP-OLS NTL data with enhanced vegetation index and land cover data. In addition, other parameters such as weather data, topography data, and agricultural land use may be useful to complement economic analysis \cite{donaldson2016view}.

\section{Data and Methods}

\subsection{Data \& Preprocessing}

We used the VNP46A2 dataset from NASA's Black Marble Product Suite that records BRDF-corrected nighttime lights (NTL) from the Day/Night Band onboard the VIIRS instrument \cite{roman2021black} to construct urban NTL time series over a period of 6 years (2013 - 2019). For each region of study, we access the time series record from February 15, 2013 up to October 21, 2019 from NASA's Level 1 and Atmosphere Archive and Distribution System Web Interface (LAADS-DAAC). Subsequently, we follow the data preprocessing methodology of \cite{chakraborty2023adaptive} to derive the gap-filled NTL time series for all pixels in urban areas corresponding to each study region, ensuring that the pixel sample contributing to the mean urban radiance of the city remains consistent. 

We focus on four hyperdense cities of the Global South, including Delhi (India), Guangzhou (China), Cairo (Egypt), and São Paulo (Brazil). The urban boundaries of our study areas are defined using the Joint Research Commission Global Human Settlements Functional Urban Areas dataset \cite{moreno2021metropolitan}. For all pixels in the urban areas of interest, we mask out pixels with a gap-filled NTL that matches the VNP46A2 fill value at each time step. Next, for the remaining pixels, we apply the DNB scale factor and sum the NTL radiance measurements weighted by each pixel's area and divide by the total estimated area of all utilised pixels to derive a radiance intensity in units of  nW/cm²/sr (nanoWatts per square centimetre per steradian). This produces the area-weighted NTL at each time step. We also extract the “Mandatory Quality Flag” from the VNP46A2 dataset to compute the ``daily gap-filled fraction'' indicating the fraction of low-quality pixels in the urban area that was gap-filled using past good retrievals. Further specifications are detailed in our replication code.

Using ArcGIS Pro 3.4, customised colour ramps are implemented to distinguish between different intensity levels of artificial light emissions. High radiance values, often associated with urban centres, industrial zones, and infrastructure corridors, are generally represented in warm colours (yellow to red), while lower radiance values, indicative of rural or undeveloped regions, are shown in cooler shades (blue to green) \cite{zheng2023nighttime}.  The spatial and temporal distribution of the exploratory NTL intensities is shown in Figure 1. And it is apparent to visually observe the increasing trend in the NTL radiance intensity from 2013 to 2019 for the selected cities.

\begin{figure}[!ht]
    \centering
    \includegraphics[width=0.9\linewidth]{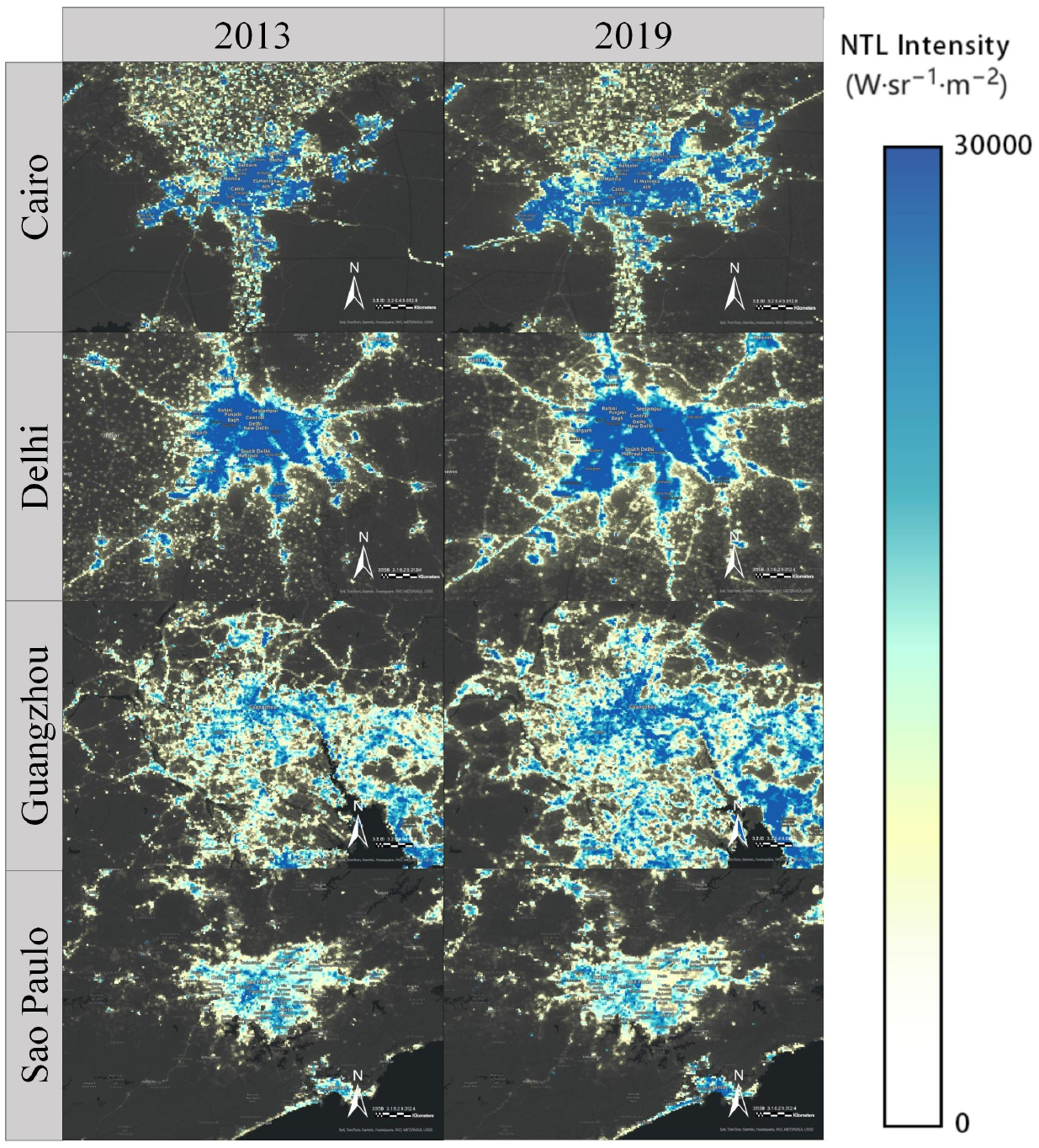}
    \caption{Visual representation of the annual mean nightime light (NTL) intensity ((in nW/cm²/sr)) for the study areas for 2013 and 2019}
    \label{fig:Figure 1}
\end{figure}

Historical daily weather and climate data used in this study were obtained from the Timeline Weather API \cite{VisualCrossing2025} for the four cities. The variables queried are listed in Table 1.

\subsection{Conceptualization}

We conceptualised a direct relationship between heatwaves and increased economic activities at night illustrated in Figure 2 as a directed acyclic graph (DAG). We use nightlight (NTL) intensity values (in nW/cm²/sr) as a proxy for the representation of urban nighttime activities (after \cite{zheng2023nighttime}), which in turn is often used as an indicator of urban economic activities \cite{chakraborty2023adaptive}. In our DAG, we assume a direct impact of heatwaves on night-time activities, denoted by increases in NTL radiance values between 2013 and 2019. In addition, we add the effects of confounding factors associated with heatwaves comprising various climate variables, lag variables, and interaction terms, as illustrated in Table 1. 

\begin{figure}[!ht]
    \centering
    \includegraphics[width=0.8\linewidth]{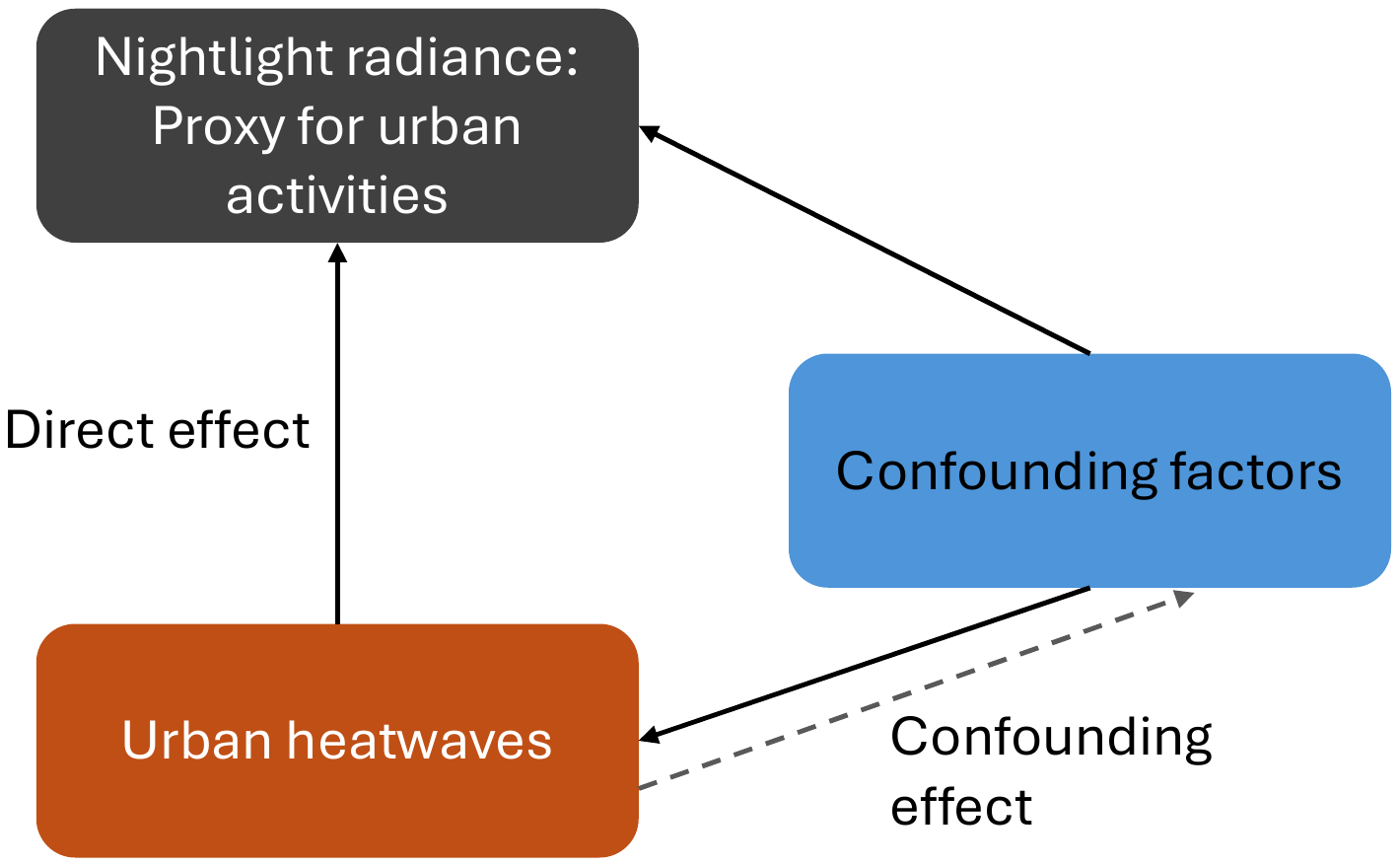}
    \caption{A directed acyclic graph (DAG) representing the causal relationship between urban heatwaves and nightlight radiance as a proxy for urban activities.}
    \label{fig:Figure 2}
\end{figure}

\begin{table}[!h]
\caption{Confounding variables used in this study}%%%Table caption goes here
\label{table1}
\begin{tabular}{|p{3cm}|p{5cm}|p{8cm}|}%%%The number of columns has to be defined here
\hline
Category &Variables &Descriptions \\
\hline
Climate variables & Cooling Degree Days (CDD) &A measure of how much (in \degree C), and for how long (in days), outside air temperature was higher than a specific base temperature.  \\
&Temp(max)  & Maximum recorded temperature in a day (in \degree C) \\
&Humidity &The amount of moisture in the air, expressed as percentage (\%) \\
&Dew &The temperature (\degree C) at which air becomes saturated with moisture \\
&Cloud cover & The fraction of the sky covered by clouds, measured in percentage (\%) \\
&Precipitation & The amount of rainfall or snowfall (in mm) recorded per day \\
&Wind speed & Average wind speed for the day measured in km/h \\
&Solar energy & Total solar radiation received during the day measured in kwh/m$^2$ \\ \hline
Lag variables & \tiny \[CDD\_Lag\{i\}, \{i\} \in (1,2,3)\] &CDD lagged by \textit{i} days, representing the heat intensity from previous days to capture delayed effects  \\
& \tiny \[Humidity\_lag\{i\}, \{i\} \in (1)\] &Humidity from the previous day, accounting for the impact of past moisture levels \\
& \tiny \[Temp\_lag\{i\}, \{i\} \in (1,2)\] &Average temperature lagged by \textit{i} days, capturing delayed effects of temperature variations \\ \hline
Interaction terms & Heatwave - humidity &Interaction term between heatwave events  and humidity, capturing how humidity modifies the effect of heatwaves \\
& Heatwave - solar energy &Interaction between heatwave events  and solar energy assessing how solar radiation changes during heatwave conditions \\
& Temp (max) - cloud cover &Interaction between maximum temperature and cloud cover, measuring how cloudy days affect the impact of high temperatures \\ \hline
\end{tabular}
\vspace*{-4pt}
\end{table}%%%End of the table

\subsection{Methodology}

In empirical research, traditional parametric methods, such as ordinary least squares (OLS), fixed effects models, and instrumental variables (IV) approaches, are widely used to estimate causal relationships. These methods typically rely on strong functional form assumptions (e.g., linear relationships) and a relatively low-dimensional set of predictors.  Although these assumptions can simplify interpretation, they become problematic when dealing with highly complex nonlinear processes or a large number of confounders ( \cite{belloni2011inference, belloni2014inference, belloni2017program, angrist2009mostly, imbens2009recent}). Many methods assume a "sparse" structure, implying that only a small number of covariates significantly affect the outcome, even if a large number are available (\cite{athey2018impact}). 

A key objective in much of this causal inference literature is average treatment effects (ATE) under the unconfoundedness assumption. Under this assumption, ATE can be identified by conditioning on observed covariates, ensuring that treatment assignment is independent of potential outcomes given these controls.  In traditional econometrics, researchers handle unconfoundedness primarily through explicit modelling and adjustment techniques (\cite{rosenbaum1983central, imbens2004nonparametric}). However, these approaches often struggle with high-dimensional confounders and complex functional relationships. In our context of evaluating the average treatment effect of heatwaves on nighttime light (NTL) emissions, the sheer variety of climatic, infrastructural, and socioeconomic factors can render a purely parametric specification vulnerable to misspecification potentially biasing causal estimates. 

Recent advances in machine learning-based causal inference methods, especially double machine learning (DML), provide a flexible, nonparametric approach to modelling confounders while preserving valid inference for causal parameters (\cite{varian2014big, mullainathan2017machine, athey2017beyond, nachtigall2025built}). In this study, we use this approach with the detailed methodological specifications presented below. 
 
\subsubsection{Modeling approach}

Firstly, we start our analysis by defining the conditions for heat waves using a two-stage approach.

\paragraph{Hot Day Identification}
For each city $c$, daily average temperatures $T_{c,t}$ were compared with a percentage-based threshold. Let $\tau_{c,p}$ denote the temperature corresponding to the $p$-th percentile of the city’s historical daily temperature distributions. The hot day indicator function is then defined as 

\begin{equation}
    \text{$I$}_{c,t} =
    \begin{cases}
        1, & \text{if } T_{c,t} \geq \tau_{c,p}, \\
        0, & \text{otherwise}.
    \end{cases}
\end{equation}
In our baseline specification, we set $p = 80$, implying  $\tau_{c,0.80}$ is the $80$-th percentile temperature for a city $c$. However, to test robustness, we also evaluate $p$ $\in$ $\{0.80,0.85, 0.90\}$. Each city’s threshold is estimated independently, reflecting differences in local climates.

\paragraph{Heatwave Period} 
A heatwave was defined as $d$ consecutive hot days. For duration  $d$, the heatwave indicator was:
\begin{equation}
    \text{HW}_{c,t}^{(p,d)} =
    \begin{cases} 
        1, & \text{if } \sum\limits_{k=0}^{d-1} \text{$I$}_{c,t-k} \geq d, \\
        0, & \text{otherwise}.
    \end{cases}
    \label{eq:binary}
\end{equation}

Thus, $HW_{c,t}^{(p,d)} = 1$ whenever the temperature in the city $c$ has exceeded its $p$-th percentile threshold for $d$ consecutive days. We apply this rolling-sum logic for different durations $d$ $\in$  $\{2,3,4\}$ to assess how longer heatwaves affect outcomes differently than shorter ones. Since each city $c$ may have different local climatic conditions, the percentile threshold $\tau_{c,p}$ is estimated separately per city, to ensure that a 'hot day'' in Delhi vs. Cairo is measured relative to its historical local distribution.

This two-stage approach, first identifying days at or above the percentile threshold and then aggregating consecutive hot days, provides flexibility in capturing varying intensities and durations of heat stress in different climate regimes.

\subsubsection*{Double machine learning approach}

Now, we estimate the effect of these heatwave events on NTL using a Double/Debiased Machine Learning (DML) framework. For each combination of ($p,d$), we define $Y_{c,t}$ as the nighttime light 
($\log(NTL_t)$) intensity (our outcome of interest in radiance), $\text{HW}_{c,t}^{(p,d)}$ as the binary treatment (heatwave indicator) and X as a set of covariates.  Following \cite{chernozhukov2018double}, we consider the partially linear regression model \footnote{DML allows the use of variety of different predictive or ML methods for the treatment and outcome model (\cite{chernozhukov2018double}). Depending on the empirical setting, one might use different ML algorithms (e.g. gradient boosting, neural networks) in place of Lasso or Random Forest. The key requirement is that each nuisance model  $\hat{g}(\cdot)$ and  $\hat{m}(\cdot)$ should be sufficiently flexible to capture confounding relationships without imposing overly restrictive functional forms}:

\begin{equation}
    Y_{c,t} = \theta^{(p,d)} \cdot \textbf{HW}_{c,t}^{(p,d)} + g(\mathbf{X}_{c,t}) + \epsilon_{c,t},
\end{equation}
\begin{equation}
    \textbf{HW}_{c,t}^{(p,d)} = m(\mathbf{X}_{c,t}) + \nu_{c,t}.
\end{equation}

where $g(.)$ and $m(.)$ are unknown nuisance functions, $\epsilon_{c,t}$ and $v_{c,t}$ the error terms, and $\theta^{(p,d)}$ denote the causal average treatment effect (ATE) of experiencing a heatwave at threshold $p$ and duration $d$. 

In the first stage, we use flexible machine learning algorithms to estimate the nuisance functions $g(.)$ and $m(.)$. In the second stage, we implement an  orthogonalized, or “double-robust,” procedure to isolate $\theta^{(p,d)}$ while mitigating bias from potential misspecification. By splitting the data into folds (cross-fitting), we avoid using the same observations for both training and final estimation of nuisance functions, thereby reducing overfitting. 

\paragraph{Nuisance function estimation} 

To operationalise DML, we must estimate the two “nuisance” components: \textbf{(i)} the outcome regression function $g(.)$ and \textbf{(ii)} the treatment assignment function $m(.)$. 
\begin{enumerate}
    \item \textbf{Outcome model} ($g$): Estimated using \textbf{Lasso regression}: 
    \begin{equation}
    \hat{g}(\mathbf{X}_{c,t}) = \arg\min_{g} 
    \left\{ \frac{1}{n} \sum_{c,t} \left( Y_{c,t} - g(\mathbf{X}_{c,t}) \right)^2 
    + \lambda \sum_{j=1}^{J} |\beta_j| \right\},
\end{equation}
where $\lambda$ is the regularization hyperparameter (often chosen via cross-validation), and $\beta_{j}$ are the regression coefficients on each feature $j$, and $n$ is the total number of observations. By shrinking certain $\beta_{j}$ towards zero,  Lasso simultaneously performs variable selection and regularised estimation, allowing $g(.)$ to capture high-dimensional or potentially sparse relationships.

    \item  \textbf{Treatment model} ($m$): Estimated using \textbf{Random Forest}: 
    \begin{equation}
    \hat{m}(\mathbf{X}_{c,t}) = \frac{1}{B} \sum_{b=1}^{B} \text{Tree}_b (\mathbf{X}_{c,t}),
\end{equation}
where each $Tree_{b}(\cdot)$ is grown on a bootstrap sample of the data (with random feature splits), and $B$ is the total number of trees in the ensemble. This approach allows for non-linear and interaction effects among covariates when predicting $\textbf{HW}_{c,t}^{(p,d)}$.   
\end{enumerate}

\paragraph{Orthogonalization and cross-fitting}

DML proposed by \cite{chernozhukov2018double} leverages orthogonalised cross-fitting, making it less susceptible to overfitting and model misspecification than naive machine learning regressions. Since semiparametric methods rely on model quality for nuisance parameter estimation, critics point out that they can be sensitive to the inclusion of irrelevant or endogenous variables, potentially inflating variance and biasing estimates (\cite{knaus2022double}). Similarly, the performance of DML is highly dependent on the correct specification of the machine learning models used for the estimation of nuisance parameters (\cite{chernozhukov2018double}). \cite{fuhr2024estimating} highlights that while DML can adjust for various non-linear confounding relationships, it still critically depends on standard assumptions about causal structure and identification. Therefore, careful consideration is necessary when selecting control variables and specifying models within the DML framework to ensure valid causal inference. 

Cross-fitting is central to DML because it ensures that no observation is used to train the nuisance functions that are then applied to the same observation in the final stage. This protects against overfitting and yields more reliable standard errors. Concretely:
\begin{enumerate}
    \item \textbf{Data Splitting} Divide the sample into $K$ folds (\textit{i.e.} $K=10$ in our baseline setting). Denote each fold by $\mathcal{I}_k$.
    \item \textbf{Training and Prediction} For each fold $k$: 
    \begin{itemize}
        \item \text{Train} the nuisance models $\hat{g}^{(k)}$ \text{and} $\hat{m}^{(k)}$ \text{on} all folds except $\mathcal{I}_k$.  
        \item Predict on the held-out fold $\mathcal{I}_k$, obtaining residuals: 
        \begin{equation}
    \tilde{Y}_{c,t}^{(k)} = Y_{c,t} - \hat{g}^{(k)}(\mathbf{X}_{c,t}),
\end{equation}
\begin{equation}
    \tilde{D}_{c,t}^{(k)} = \mathbf{HW}_{c,t}^{(p,d)} - \hat{m}^{(k)}(\mathbf{X}_{c,t}).
\end{equation}
Here $\tilde{Y}_{c,t}^{(k)}$ is the outcome “partialed out” by the predicted outcome model, and $\tilde{D}_{c,t}^{(k)}$ is the treatment “partialed out” by the predicted treatment model.  
    \end{itemize}
    \item \textbf{Final estimation of $\theta^{(p,d)}$}:  After obtaining all residuals $\tilde{Y}_{c,t}^{(k)}$ and $\tilde{D}_{c,t}^{(k)}$ from each fold, DML stacks them together and run a simple linear regression (or direct method-of-moments formula): 
    \begin{equation}
    \hat{\theta}^{(p,d)} = 
    \left( \sum_{c,t} \tilde{D}_{c,t}^{2} \right)^{-1} 
    \left( \sum_{c,t} \tilde{D}_{c,t} \tilde{Y}_{c,t} \right).
\end{equation}
This step effectively regresses the “partialed-out outcome” $\tilde{Y}_{c,t}^{(k)}$ to the “partialed-out treatment” $\tilde{D}_{c,t}^{(k)}$. Because the nuisance models were trained on different folds than the ones used for the final regression, the resulting estimator $\hat{\theta}^{(p,d)}$ is orthogonal (\textit{i.e. "doubly robust"}) to small errors in $\hat{g}$ and $\hat{m}$. 
\end{enumerate}

\paragraph{Estimation approach by Lasso and Random Forest}

Lasso is a penalised regression method designed to handle highly dimensional or potentially sparse data. Lasso and tree-based methods, such as random forests and boosted trees, can perform variable selection by default. That is, they will only make use of features that are predictive of the outcome,  which enables them to work in high-dimensional settings with many variables or transformations. In our study, we implement lasso with the glmnet package (v 4.1) and random forest with the ranger package (v 0.7.0) with default settings in R.

%\paragraph{Conditional treatment effects}
%Given average treatment effect (ATE) captures the overall impact of heatwaves on nighttime light (NTL), in many instances we are also interested in how this effect varies across different values of the covariates \mathbf{X}. Formally, we define the conditional average treatment effect (CATE) at a point 
%\math{x}: 

%\[
%\tau(x) = \mathbb{E} [ Y(1) - Y(0) \mid X = x ],
%\]

%where $Y(1)$ and $Y(0)$ are potential outcomes under treatment and control, respectively. To estimate CATE, we consider the DR-learner under the DML framework which enables the estimation and construction of confidence intervals for CATEs.

%\begin{itemize}
%    \item Estimate Nuisance Function
%   \item Form Residuals:
%    \[
%\tilde{Y}_i = Y_i - \hat{g}^{(-k_i)}(X_i), \quad
%\tilde{D}_i = D_i - \hat{m}^{(-k_i)}(X_i),
%\]
%where \mathbf{D_{i}} $\in \{0,1\}$ indicates absence/presence of a heatwave. These residuals are computed in a cross-fitted manner. 

%\item Pseudo-Outcome (DR-Learner) for Heterogeneity
%\[
%\hat{\phi}_i = \left( \tilde{Y}_i \right) \left( \tilde{D}_i \right)
%\]
%where $\hat{\phi}_i$ approximates the individual-level contribution to the treatment effect.

%\item Second-Stage Regression: Regress $\hat{\phi}_i$ on $\mathbf{X_{i}}$ using a flexible method (e.g.,  random forest, splines, or OLS). The fitted function $\hat{\tau}(x)$ is then the estimated conditional treatment effect at $\math{x}.$

%\end{itemize}

\paragraph{Model assumptions}
We follow \cite{nachtigall2025built}'s specifications to obtain a reliable treatment estimate from observational data using causal inference methods. Several assumptions must be met, most importantly unconfoundedness, ignorability, positivity, and the Stable Unit Treatment Value Assumption (SUTVA). Unconfoundedness asserts that covariates encapsulate all influencing factors on both treatment and outcome, leaving no confounders unaccounted for, thus removing confounding bias.

Ignorability asserts that, when conditioned on observed covariates, the assignment of treatment is autonomous of potential outcomes. As such, units could be swapped between the treatment group (presence of a heatwave event) and control group (absence of heatwave event) without affecting the experiment (exchangeability). Incorporating inappropriate controls, such as variables that can influence heatwave estimations like urban heat island effect or specific definition of heatwaves used by the region, breaches the assumption and skews the effect estimate. Therefore, in our study, we use cooling degree days (CDD) to account for the region-specific definition effect.

Positivity states that, across the range of covariates, there must be a positive probability that observations are present in the treatment group. In other words, there should be a reasonable overlap between the covariate distribution of the treatment and control groups. For example, heatwaves increase cooling degree days (CDD). Lastly, SUTVA states that assigning treatment to one unit does not affect the potential outcomes of other units. It assumes that there are no interference or spillover effects between units. It also states that there is only one consistent version of treatment. In our case, the climatic variables used are physical phenomena and do not interfere with NTL activities. We introduce lag variables to capture these spillover effects (see Table 1). For the moment, we assume that these assumptions hold sufficiently and discuss potential violations and implications for the robustness of our results and the applicability of the model in the discussion.

\subsubsection*{Evaluation}

To facilitate the interpretation of the treatment effect, we report it as a log of average nighttime light (NTL) radiance compared to heatwave days across each city. As our unit of analysis is the heatwave events, we report the absolute individual treatment effect (ATE) in the direction of more nighttime activities due to heatwaves, i.e., higher heatwave events and greater nightlight radiance. Here, ATE gives an indicator of how much the existing difference in heatwave events affects nighttime activities in Delhi, Guangzhou, Cairo, and São Paulo. 

To examine the robustness of our results, we performed the following experiments: (1) We assess the existence of threshold effects and other nonlinearities of the influence of the heatwave on NTL in the study areas using an XGBoost model and discuss their implications for our DML approach. (2) We performed a sensitivity analysis to assess the variations in the NTL values before and after the heatwave analysis. Here, we considered the start of the heatwave event as the baseline (0th day), and 2 days before and 5 days after this baseline event. Furthermore, we address the low explanatory power and the resulting high uncertainty of treatment effect estimates by performing lag event analysis by 1, 2, and 3 days, respectively, across temperature, humidity, and cooling degree-day confounders (see Table 1). Detailed robustness check results are presented in SI A5 and A6.

\section{Results}

We structure the results into three parts. The first is descriptive statistics exploring the relationship between heatwave events and nighttime light (NTL) activities. Second, we expand this relationship based on our heatwave episode thresholds and report the percentage change in NTL activities during heatwave events in the four cities. Finally, in the third part, we report the results of the double machine learning analysis, its inference, and discuss its robustness using a sensitivity analysis. 

To explore whether there is a relationship between daily air temperature and nighttime light (NTL) activities in the four cities, we use the Granger Causality Test. We find statistically significant associations (99\% CI) between air temperature and NTL activities in all cities. This suggests that elevated air temperatures commonly observed during heatwave episodes may lead to higher NTL radiance values, a proxy for increased nighttime activities. In addition, we performed the Augmented Dickey-Fuller (ADF) test for all cities, which confirmed stationarity for NTL and air temperature time series data in the study areas. Extended descriptive results are presented in the Supplementary Information (SI) (see SI A1 - A3). 

\subsection{Threshold selection}

City-specific temperature thresholds (80th - 90th percentiles) were chosen through sensitivity testing because each city’s baseline climate differs substantially. Initially, we experimented with multiple percentile thresholds ($80\%$, $85\%$, $90\%$) to identify which one reflected both local notions of `extreme heat' and yielded stable estimation results. (\text{i.e.} manageable standard errors, robust inference). A threshold of 90\% in São Paulo may capture a distinctly high (but locally relevant) temperature level, while in Guangzhou a 80\% cutoff point is already sufficiently high to define `extreme' in that regional climate. Defining a common percentile for all cities could misclassify more moderate warm events as `heatwaves' in cooler regions or fail to capture extreme events in hotter ones.  Using city-specific thresholds, we ensure that the `heatwave' label reflects a truly unusual temperature event in that context. Thus, the final cut-off points of 80\% for Guangzhou \& Cairo, 85\% for Delhi, and 90\% for São Paulo best reflect the local climate profile of each city while maintaining reasonable sample sizes for inference. We present the threshold profiles for all cities in Table 2 (standard errors are presented in SI A4). 

\begin{table}[!ht]
\caption{Heatwave threshold selection criteria for each city to emphasise the effect of local climatic profile on NTL values}%%%Table caption goes here
\label{table_example}
\begin{tabular}{|p{2cm}|p{3cm}|p{3cm}|p{2cm}|p{3cm}|p{2cm}|}%%%The number of columns has to be defined here
\hline
City&Threshold (percentile)&Optimal duration (days)&NTL increase (\%)&Standard error& p-values \\
\hline
São Paulo& 90th&  4& 59.44& 0.1486 &0.0016***\\
Delhi& 85th& 3& 14.04& 0.0564&0.0198**\\
Cairo& 80th& 3& 263.42& 0.4966&0.0093*** \\
Guangzhou& 80th& 3& 72.95& 0.2384&0.0215**\\ \hline
\end{tabular}
{\textsuperscript{***}$p<0.01$, 
  \textsuperscript{**}$p<0.05$, 
  \textsuperscript{*}$p<0.1$}
\end{table}%%%End of the table

To ensure empirical robustness, we examined how each threshold influenced the statistical precision (standard errors, p-values) of our double machine learning estimates. Some thresholds led to very small effective sample sizes driving high variance, while others were more reliably aligned with the observed changes in night-time light (NTL). Ultimately, we selected the percentile threshold that yields the best trade-off between (a) capturing genuine extreme heat episodes and (b) providing stable, consistent estimates. Detailed robustness check results are presented in SI A5 and A6. 

\subsection{Direct relationships}

\begin{figure}[!ht]
     \centering
     \begin{subfigure}[b]{0.47\textwidth}
         \centering
         \includegraphics[trim=0cm 0.6cm 0cm 0.6cm,clip,width=\textwidth]{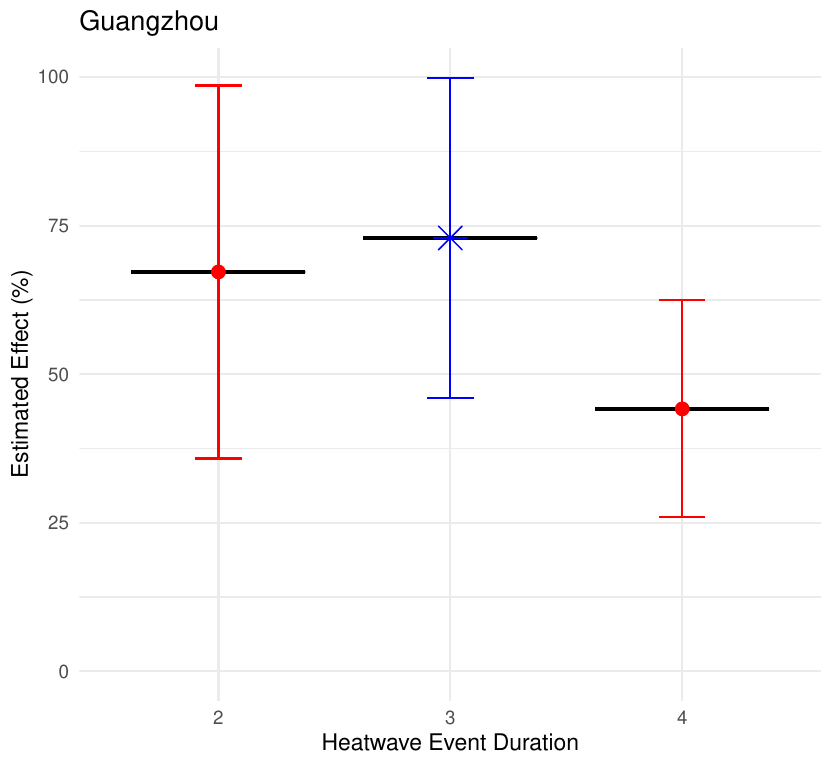}
         \caption{Guangzhou}
         \label{fig:Fig3a}
     \end{subfigure}
     \hfill
     \begin{subfigure}[b]{0.47\textwidth}
         \centering
         \includegraphics[trim=0cm 0.6cm 0cm 0.6cm,clip, width=\textwidth]{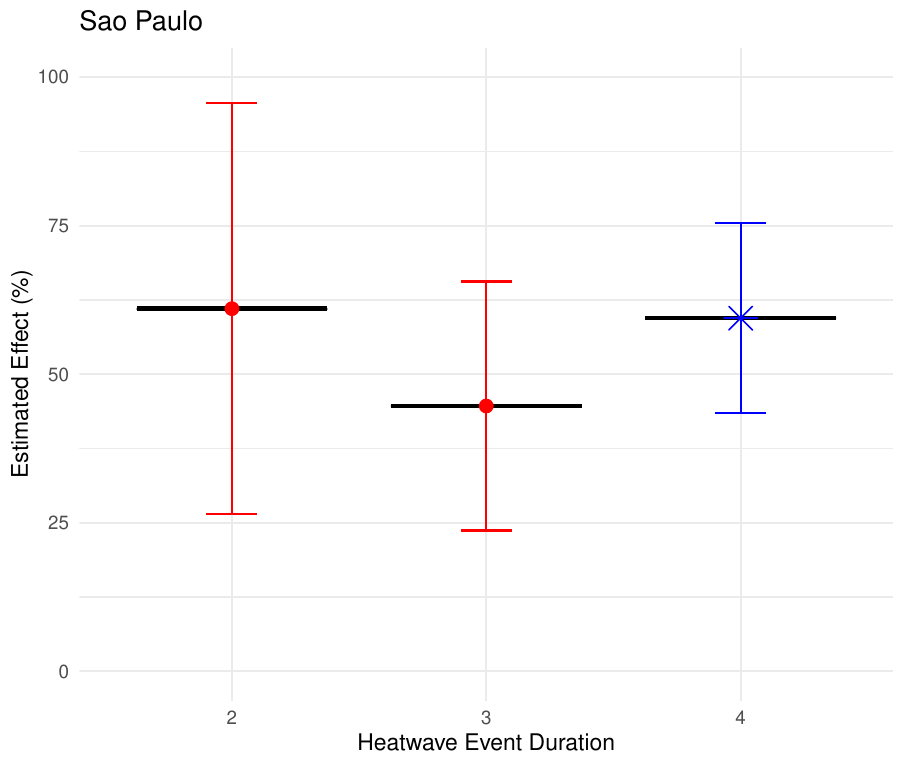}
         \caption{São Paulo}
         \label{fig:Fig 3b}
     \end{subfigure}
     \hfill
     \begin{subfigure}[b]{0.47\textwidth}
         \centering
         \includegraphics[trim=0cm 0.6cm 0cm 0.6cm,clip, width=\textwidth]{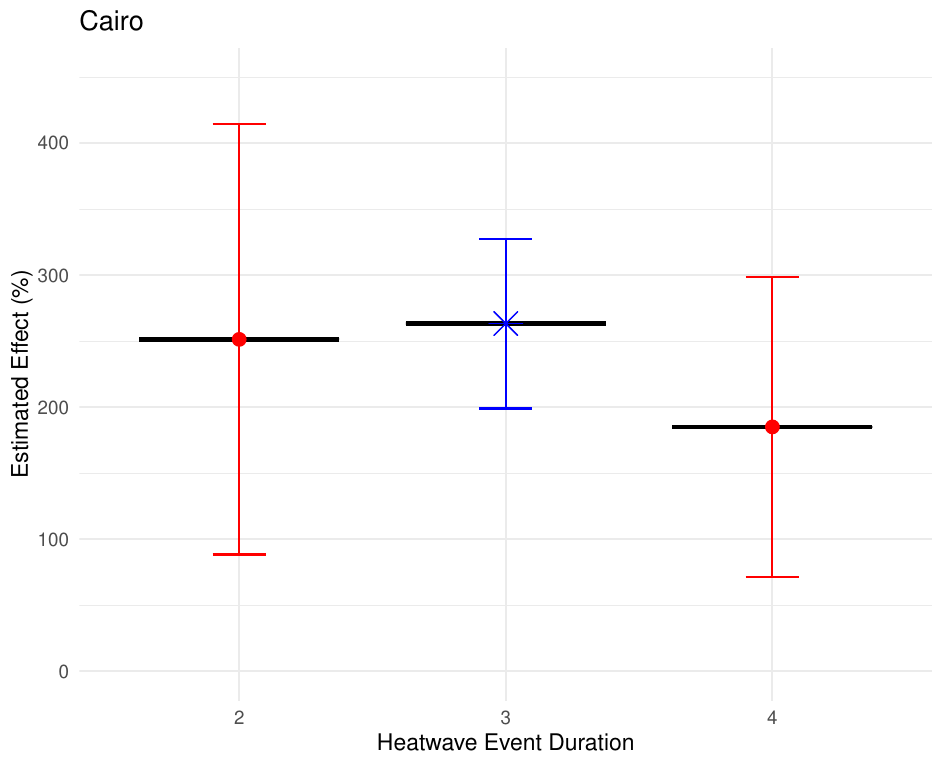}
         \caption{Cairo}
         \label{fig: Fig3c}
     \end{subfigure}
          \begin{subfigure}[b]{0.47\textwidth}
         \centering
         \includegraphics[trim=0cm 0.6cm 0cm 0.6cm,clip,width=\textwidth]{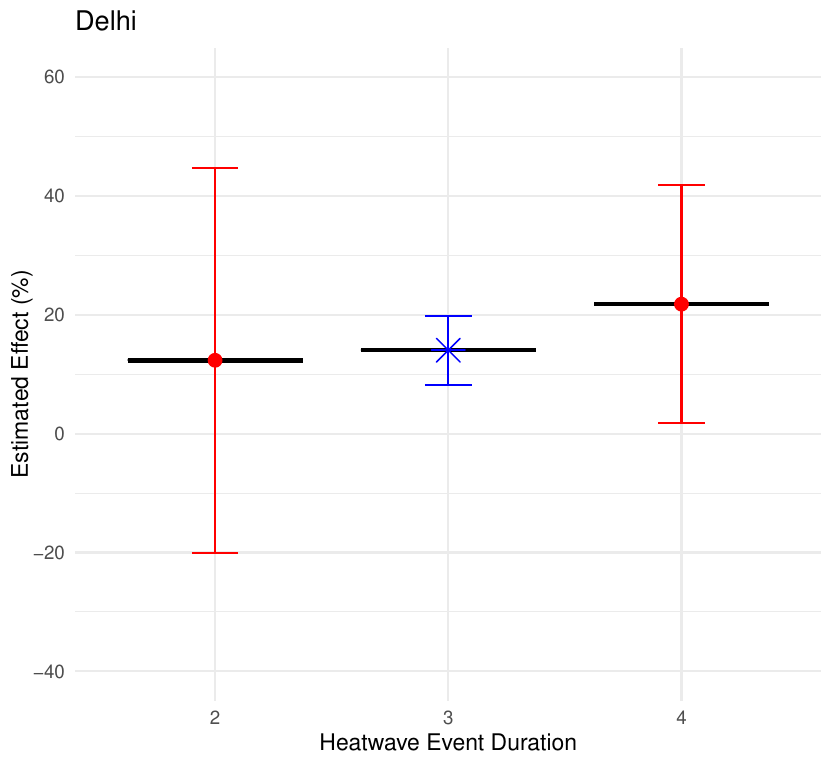}
         \caption{Delhi}
         \label{fig:Fig3d}
     \end{subfigure}
     \hfill
        \caption{Average treatment effect (ATE) results demonstrating the effect of heatwave events on urban nightime activities in four study areas, denoted by nighttime light (NTL) radiance values. The x-axis denotes the heatwave episode on the day 2, day 3 and day 4, respectively and y-axis shows the estimated ATE (in \%). Error bars shown at 95\% CI.}
        \label{fig:3}
\end{figure}

In this study, we used the double machine learning (DML) method to investigate how heatwaves affect urban nighttime activities within four hyperdense cities located in the Global South. Our findings indicate that during a heatwave, there is a notable effect on the nighttime light (NTL) values on the third day (Day 3 in the cities of Delhi, Cairo and Guangzhou. This observation is quantitatively represented as the average treatment effect (ATE) in percentage (\%) in Figure 3 with 95\% CI. However, in the case of São Paulo, a significant impact on the NTL values is observed on the fourth day (Day 4) of the heatwave. This discrepancy can be attributed to variations in climatic conditions, temperature ranges, and unique urban characteristics inherent to these metropolitan areas \cite{levin2020remote}.

In all cities, heatwaves (defined as consecutive days that exceed city-specific temperature thresholds) are associated with increased NTL, although the magnitudes of the effect and statistical significance vary. Longer heatwaves often amplify the effects, but the optimal duration differs by city (São Paulo shows the strongest effects at $d =4$ while others peak at $d=3$). However, certain heatwave events yield large but imprecise estimates that point to significant heterogeneity in how consecutive hot days impact local nighttime activity. Heatwaves of $3+ \text{or } 4+$ consecutive days produce the most robust and significant results in the study areas. This implies that a single hot day may not drastically change behaviour or urban activity, but prolonged exposure triggers more noticeable responses.  When daytime heat becomes oppressive, especially for more than a day, residents adapt by shifting certain tasks (e.g., shopping, social gatherings, outdoor errands) to cooler evening hours, thus increasing electricity and lighting usage (according to \cite{graff2014temperature}). 

\begin{figure}[h!]
     \centering
     \begin{subfigure}[b]{0.47\textwidth}
         \centering
         \includegraphics[width=\textwidth]{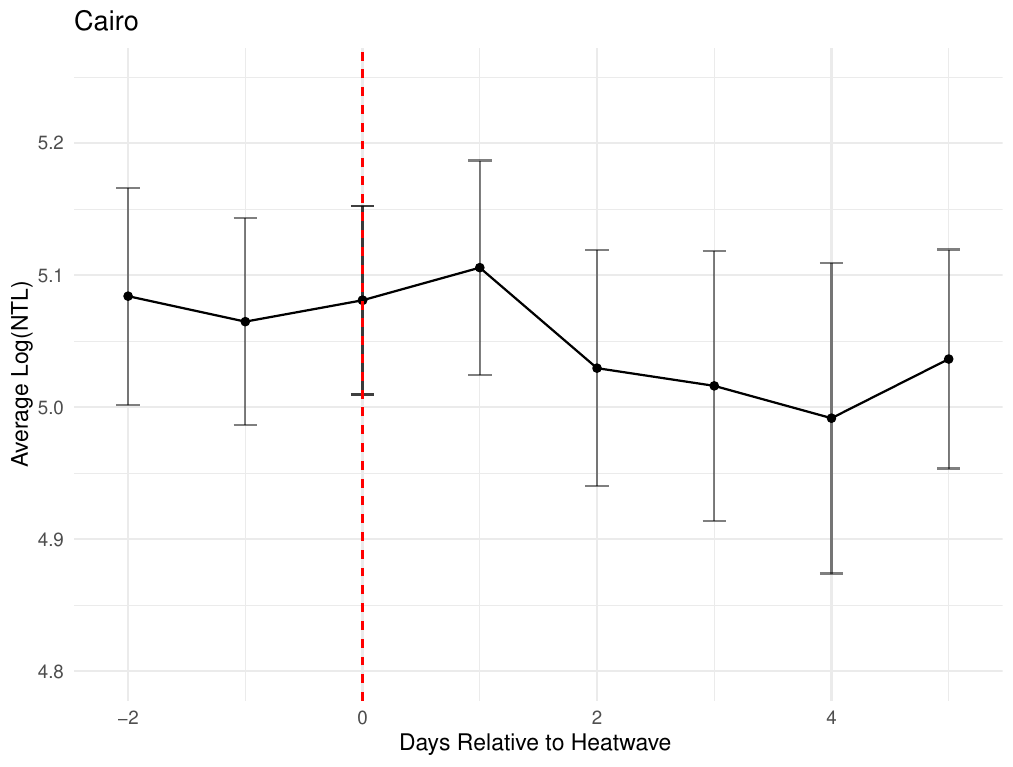}
         \caption{Cairo}
         \label{fig:y equals x}
     \end{subfigure}
     \hfill
     \begin{subfigure}[b]{0.47\textwidth}
         \centering
         \includegraphics[width=\textwidth]{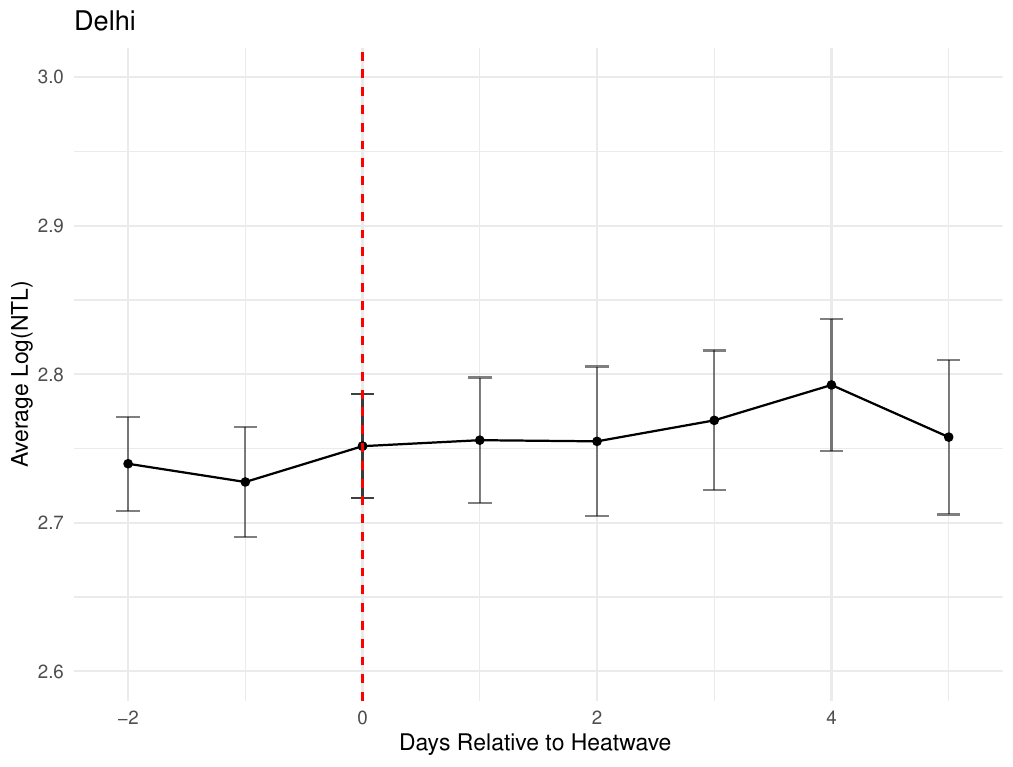}
         \caption{Delhi}
         \label{fig:three sin x}
     \end{subfigure}
     \hfill
     \begin{subfigure}[b]{0.47\textwidth}
         \centering
         \includegraphics[width=\textwidth]{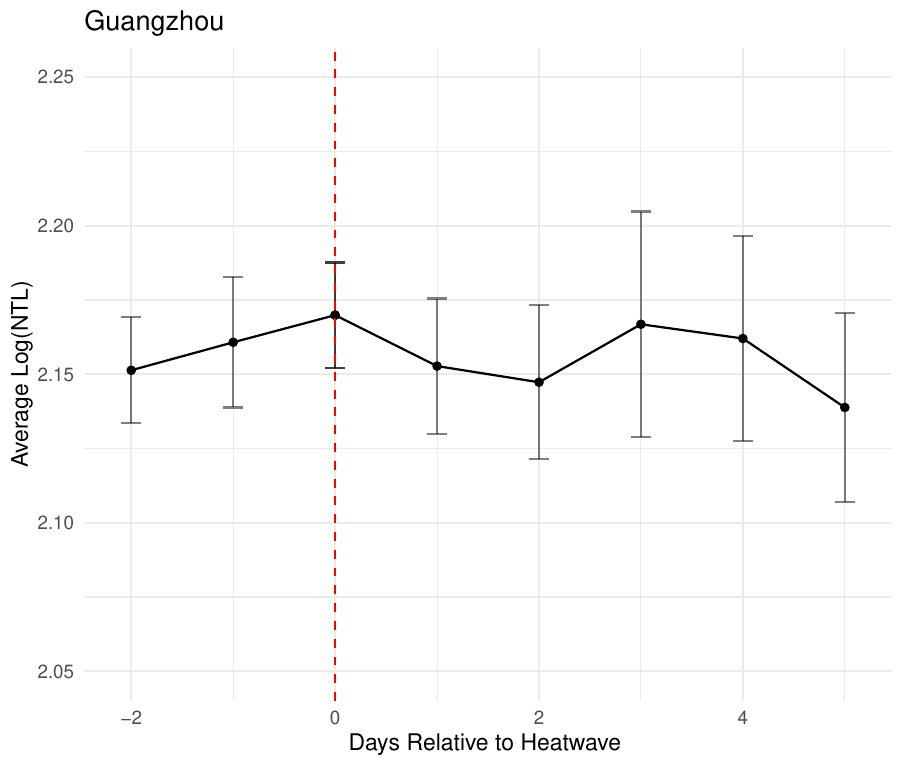}
         \caption{Guangzhou}
         \label{Guangzhou}
     \end{subfigure}
          \begin{subfigure}[b]{0.47\textwidth}
         \centering
         \includegraphics[width=\textwidth]{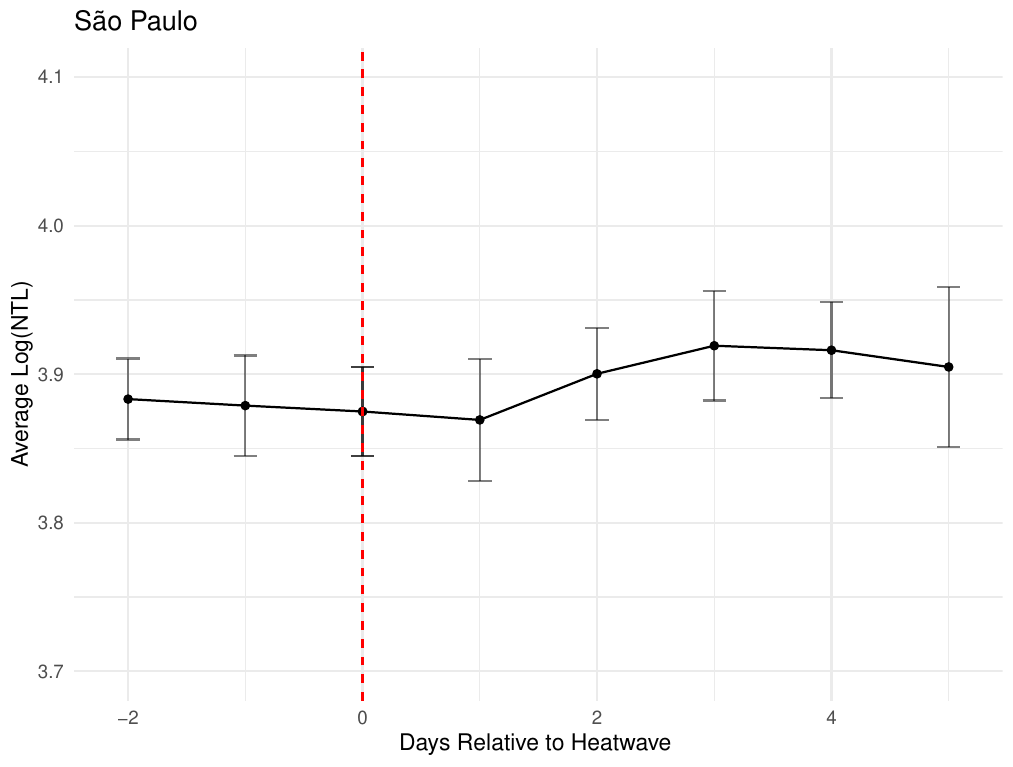}
         \caption{São Paulo}
         \label{fig:five over x}
     \end{subfigure}
     \hfill
        \caption{Sensitivity analysis illustrating before and after impact impacts of heatwave events (in days) on nighttime light (NTL) radiance across the four cities. Error bars shown at 95\% CI.}
        \label{fig:three graphs}
\end{figure}

A close examination of São Paulo and Guangzhou reveals notable similarities in how consecutive hot days influence NTL values, despite differing average climates and temperature thresholds. In Guangzhou, we identified 294 days of heatwaves below the 80th percentile threshold for 3 + consecutive hot days and 220 such days for 4+ consecutive hot days. In contrast, São Paulo shows 92 heatwave days at the 90th-percentile threshold for 3+ days and 63 for 4+ days. We opt for a higher percentile ($90\%$) in São Paulo given its lower average temperature, to ensure that we capture days truly characterised by extreme local heat. Despite these numerical differences, both cities show consistent and moderate increases in NTL intensity values across the chosen thresholds once we move to 3+ or 4+ consecutive hot days (see Figure 4). This suggests that while Guangzhou experiences more frequent heatwave events (due to an overall warmer baseline climate), São Paulo responds similarly once temperatures surpass its locally meaningful `extreme' mark. 

Although both Delhi and Cairo experience frequent heat waves, their nightlight (NTL) responses diverge starkly, with Cairo exhibiting extreme NTL surge ($263\%$ increase for 3+ day heatwaves) compared to Delhi’s muted effects ($14\%$ increase for 3+ day heatwaves) (see Figure 4). There may be numerous regional heterogeneity factors that explain this contrast in NTL between Delhi and Cairo, including the use of different types of lighting together with different street lighting standards \cite{levin2020remote}. Holiday and ornamental lights, political, historical, and cultural differences in lighting can result in a temporary increase in NTL usage. For example, studies have reported such observation during major cultural events such as Christmas, New Year, Diwali, and the holy month of Ramadan \cite{roman2015holidays}; and tourist-related NTL spikes \cite{stathakis2018seasonal}. However, we cannot validate these hypotheses within the scope of this study and remain an aspiration for future work.

\section{Discussion and Conclusion}

\subsection{Heatwaves increase intensities of urban nighttime light}

In this study, we investigated the relationship between heatwave events and nightime light (NTL) from 2013 to 2019 in four hyperdense cities of the Global South using a double machine learning framework. We scoped the study based on evidence in the current literature that NTL can be used as a proxy for night activities and economic productivity; therefore, we wanted to measure the impact of heatwave events on nighttime economic productivity in urban areas. In doing so, we adjusted for local climatic conditions (see Table 1).

We observe a statistically significant direct link between heatwaves and the elevated intensity of NTL in Delhi, Guangzhou, Cairo, and São Paulo (see Figure 3 and Table 2). The results indicate that for Cairo, Delhi and Guangzhou, the intensity of NTL increases on the third day of the heatwave, while in São Paulo, a significant effect appears on the fourth day. Furthermore, a sensitivity analysis was performed to assess the robustness of these findings by evaluating changes in NTL intensity before and after the heatwave (see Figure 4). This sensitivity analysis reveals that consistent heatwave response patterns in the study areas suggest an important adaptive implication for urban populations: successive hot days may prompt residents to adjust their daytime activities to nighttime. This finding is consistent with the review of \cite{cheval2024systematic}.

At an aggregated level, the increase in NTL radiance on days of heatwave suggests that individuals and businesses can modify their schedules to avoid extreme daytime temperatures. This aligns with existing research showing that excessive heat reduces labour productivity, increases energy demands, and disrupts urban mobility \cite{somanathan2021impact, schlenker2009nonlinear, miller2021heat}. The observed time lag effect, where significant increases in NTL intensities occur on the third or fourth day of a heat wave, indicates that the shift to nighttime activity is not immediate but builds up as heat stress persists. This delay may be due to a combination of physiological adaptation, urban infrastructure limitations, and socioeconomic constraints that prevent an immediate transition to patterns of nocturnal activity.

Furthermore, the heterogeneity in heatwave responses across cities suggests that urban morphology and climate adaptation capacity influence how populations cope with rising temperatures. São Paulo, for example, exhibits a slightly delayed response (Day 4 effect), which could be attributed to differences in building materials, energy infrastructure, and urban cooling mechanisms. In contrast, cities like Cairo and Guangzhou show a more immediate increase in NTL, potentially reflecting more flexible labour structures or pre-existing economic activity patterns that allow for greater adaptability. At the population level, the diversity of individual heat adaptive capacities are different across the chosen cities, which also results in the variation of response time.  

Our results have important implications for climate adaptation and the management of the urban heat island effect, especially in rapidly urbanising cities of the Global South. As heatwaves become more frequent, there may be a greater shift of urban activities toward nighttime, presumably as an adaptive response to extreme heat exposure during the day. Consequently, sufficient resources and infrastructure are essential to facilitate these transformations. On the urban scale, for example, activity changes could alter the energy demand curve, potentially overburdening the electricity grid and risking grid failure. Similarly, the real impact of these changes on economic productivity must be measured to design policies that can support this transition without harming the economic growth aspirations of these cities.

\subsection{Policy implications for the planning of heat-resilient infrastructure}

The observed increase in NTL intensity during heatwave days suggests that as extreme temperatures increase, urban residents and businesses shift their activities to cooler nighttime hours. Although this adaptation is a natural response to increased daytime heat stress, it also underscores the urgent need for urban planning strategies that mitigate heat exposure while ensuring that such adaptations do not exacerbate energy demand, public health, and socioeconomic inequalities. A crucial policy intervention involves expanding the urban green infrastructure, as research has consistently demonstrated that vegetation plays a significant role in reducing the effect of urban heat islands (UHI). Trees, green roofs, and vertical gardens can lower surface and ambient temperatures, particularly in dense urban areas where built environments retain excessive heat (\cite{cheval2024systematic, li2024cooling}). However, these cooling benefits are often unevenly distributed, disproportionately benefiting wealthier districts while leaving lower-income communities more exposed to extreme heat. \cite{li2024cooling} highlight that trees in socially vulnerable neighbourhoods provide significantly greater cooling effects, reinforcing the need for equitable investment in green infrastructure in the most heat-affected urban areas.
 
Beyond green infrastructure, urban zoning and building codes must be adapted to enhance climate resilience. Current construction materials and urban layouts frequently increase heat retention, worsening the impact of heatwaves (\cite{chakraborty2023adaptive}). Cities can mitigate this effect by implementing climate-adaptive zoning policies that mandate the use of high-albedo materials, cool roofs, and reflective pavements in new developments. In addition, zoning regulations in such areas could be revised to incorporate ventilation corridors, open spaces that facilitate natural airflow, and reduce localised heat buildup. Research by \cite{burger2021modelling} shows that cities with well-designed wind corridors experience lower nighttime heat retention, an insight that is particularly relevant given the increasing shift toward nocturnal urban activity during extreme heat events. However, this can be challenging in a hyperdense urban development context; therefore, there must be provisions to include such urban heat-mitigation measures in rapidly urbanising cities.

In addition, district cooling systems can provide an energy-efficient alternative to air conditioning, helping to reduce spikes in energy consumption at night. These centralised cooling networks, which distribute chilled water for air conditioning purposes, have been successfully implemented in cities such as Singapore and Dubai, demonstrating their feasibility in heat-prone urban environments (\cite{buster2024tackling}). From an economic policy lens, there should be provisioning to absorb macroeconomic shocks due to productivity loss during heatwaves, which can include heat insurance and fiscal support mechanisms (as discussed by the World Bank \cite{ranger2022assessing}). 

Realising such policy frameworks in heat-resilience planning remains a future work of this study. In addition, future work aims to expand the confounding factor dataset by including sociodemographic datasets that can account for the high granularity in urban response during heatwaves. For example, specifically studying NTL in vulnerable areas and its effect during extreme heat events can help prioritise climate actions. 

\section{Limitations and future work}

This study has certain limitations related to our conceptualisation of the direct impact of heatwaves on NTL intensities in study areas between 2013 and 2019. Although the validity of NTL as a proxy for urban economic activity is well-established in literature, it is usually affected by multiple and complex confounding factors, stemming from the sociodemographic and cultural fabric of urban spaces, which are difficult to objectively specify in a data-driven model like double machine learning (DML). Similarly, different cities have different NTL regulations which also affect their measurement from space. Furthermore, remote sensing data used for NTL analysis may be subject to technical limitations, such as sensor resolution, atmospheric interference, and temporal inconsistencies in satellite data collection. These factors can affect the accuracy of NTL measurements, potentially leading to discrepancies in observed trends. Future research could benefit from integrating complementary data sources, such as high-resolution socioeconomic indicators or localised surveys, to improve the robustness of the findings. Using data sets from more recent satellites such as Jilin-1 may be more advantageous. This is because the resolution would be around 130m compared to the 500m for the VNP46A2. Future research may further study the usefulness of different satellite data.

Furthermore, in conceptualizing how the urban heat island effect (UHI) interacts with heatwaves, we adopted a simplified aggregated definition. The interactions between UHI and heat waves can be complex, especially within intricate urban structures such as density, diversity and mixed land use areas in cities. The disentanglement of these interaction effects remains an unresolved question and is a future research avenue. Likewise, we did not consider confounding factors resulting from urban morphologies in our study regions, such as the integration of shopping centres with residential areas, which may affect the intensities of the NTL. Additionally, dimly lit or dark rural regions - although they host a significant proportion of the population - are often under-represented in the models \cite{doll2010estimating}. But since our focus on this paper is urban activity and economic activity, our key signals came from high-radiance intensities (brightly lit) urban cores (industrial zones, commercial districts, city centres). In that sense, low-intensity rural lights that may be negligible are unlikely to capture the city-centric patterns we care about. Hence, our scope was limited to urban centres. 

Ongoing advances in DML frameworks and the rapid growth of causal machine learning underscore the potential for confounding errors in our models despite rigorous robustness checks and evaluations. Cross-validation using newer model alternatives such as physics-based machine learning can guide future extensions of this work.

\section*{Acknowledgments}
RB and RD thanks the UKRI ODA grant and the Minderoo Foundation for the funding support through the Lethal Humidity Global Council. The author also thanks CRASSH at Cambridge for supporting the climaTRACES lab.

\section*{Authors contribution}
R.D., T.C., F.H., and R.B. conceptualised the study. T.C. and F.H. led data preprocessing, analysis and visualisation. All authors wrote the original draft and revised the manuscript. R.D. provided overall supervision. 

\section*{Data and code availability}
All datasets used in this study is available in the public domain: i) Nightlight data using the VNP46A2 dataset from NASA's Black Marble Product Suite and ii) Timeline Weather API for daily climate data. \\

The codes used in this study are available here: \url{https://github.com/SimonRogersHan/Climate-Data}

%Bibliography
\bibliographystyle{unsrt}  
\bibliography{references}  

\newpage



\section*{Supplementary material}

# Supplementary Information (SI)

# Heatwave increases nighttime light intensity in hyperdense cities of the Global South: A double machine learning study

Ramit Debnath[1]*, Taran Chandel[1,2], Fengyuan Han[1] and Ronita Bardhan[1]

1. University of Cambridge, UK
2. University of Edinburgh, UK
*corresponding author: rd545@cam.ac.uk

## A1. Descriptive Statistics

Sao Paulo NTL Radiance Statistics:

| Min. | 1st Qu. | Median | Mean | 3rd Qu. | Max. |
|---|---|---|---|---|---|
| 0.00 | 45.64 | 48.45 | 47.69 | 50.69 | 59.91 |

Sao Paulo Air Temperature Statistics:

| Min. | 1st Qu. | Median | Mean | 3rd Qu. | Max. |
|---|---|---|---|---|---|
| 7.60 | 18.20 | 20.80 | 20.62 | 23.20 | 29.10 |

Guangzhou NTL Radiance Statistics:

| Min. | 1st Qu. | Median | Mean | 3rd Qu. | Max. |
|---|---|---|---|---|---|
| 6.009 | 8.436 | 8.851 | 8.864 | 9.345 | 12.193 |

Guangzhou Air Temperature Statistics:

| Min. | 1st Qu. | Median | Mean | 3rd Qu. | Max. |
|---|---|---|---|---|---|
| 4.40 | 18.90 | 25.10 | 23.48 | 28.40 | 33.70 |

Delhi NTL Radiance Statistics:

| Min. | 1st Qu. | Median | Mean | 3rd Qu. | Max. |
|---|---|---|---|---|---|
| 0.00 | 15.23 | 16.92 | 17.29 | 18.81 | 50.96 |

Delhi Air Temperature Statistics:

| Min. | 1st Qu. | Median | Mean | 3rd Qu. | Max. |
|---|---|---|---|---|---|
| 8.50 | 19.30 | 28.10 | 25.77 | 31.20 | 39.40 |

Cairo NTL Radiance Statistics:

| Min. | 1st Qu. | Median | Mean | 3rd Qu. | Max. |
|---|---|---|---|---|---|
| 0.0 | 119.0 | 133.0 | 141.4 | 152.2 | 349.0 |

Cairo Air Temperature Statistics:

| Min. | 1st Qu. | Median | Mean | 3rd Qu. | Max. |
|---|---|---|---|---|---|
| 7.80 | 18.40 | 24.50 | 23.58 | 28.90 | 37.90 |

## A2. ADF Test for Stationarity

| City | P-value (Air Temperature) | P-value (NTL Radiance) |
|------|---------------------------|------------------------|
| Cairo | 0.023** | <0.01*** |
| Delhi | 0.025** | <0.01*** |
| Guangzhou | 0.027** | <0.01*** |
| São Paulo | <0.01*** | <0.01*** |

***p<0.01, **p<0.05, *p<0.1

## A3. Granger Causality test confirming the relationship between heatwave events and the nightime light activities

| City | P-value |
|------|---------|
| Cairo | 1.331e-12*** |
| Delhi | 4.282e-16*** |
| Guangzhou | 3.453e-4*** |
| São Paulo | 2.2e-16*** |

***p<0.01, **p<0.05, *p<0.1

## A4. DML results

| São Paulo - DML Results for Heatwave Impact on NTL | | | | | | |
|------|------|------|------|------|------|------|
| Variable | Estimate | Std_Error | t_value | p_value | Percent_Change_NTL | Percent_Change_Se |
| is_heatwave (d=2) | 0.4765134 | 0.2973322 | 1.602630 | 0.109016444 | 61.04497 | 34.62625 |
| is_heatwave (d=3) | 0.3692083 | 0.1901942 | 1.941217 | 0.052231914 | 44.65888 | 20.94844 |
| is_heatwave (d=4) | 0.4665479 | 0.1486567 | 3.138425 | 0.001698584 | 59.44803 | 16.02746 |

### Guangzhou – DML Results for Heatwave Impact on NTL

| Variable | Estimate | Std_Error | t_value | p_value | Percent_Change_NTL | Percent_Change_Se |
|---|---|---|---|---|---|---|
| is_heatwave (d=2) | 0.5141717 | 0.2729755 | 1.883582 | 0.05962154 | 67.22527 | 31.38680 |
| is_heatwave (d=3) | 0.5478818 | 0.2384244 | 2.297926 | 0.02156600 | 72.95855 | 26.92478 |
| is_heatwave (d=4) | 0.3658853 | 0.1677668 | 2.180916 | 0.02918964 | 44.17899 | 18.26608 |

### Delhi – DML Results for Heatwave Impact on NTL

| Variable | Estimate | Std_Error | t_value | p_value | Percent_Change_NTL | Percent_Change_Se |
|---|---|---|---|---|---|---|
| is_heatwave (d=2) | 0.1164809 | 0.28072533 | 0.4149283 | 0.67819439 | 12.35360 | 32.408987 |
| is_heatwave (d=3) | 0.1314556 | 0.05643937 | 2.3291477 | 0.01985124 | 14.04873 | 5.806246 |
| is_heatwave (d=4) | 0.1972392 | 0.18254177 | 1.0805157 | 0.27991259 | 21.80354 | 20.026428 |

### Cairo – DML Results for Heatwave Impact on NTL

| Variable | Estimate | Std_Error | t_value | p_value | Percent_Change_NTL | Percent_Change_Se |
|---|---|---|---|---|---|---|
| is_heatwave (d=2) | 1.256848 | 0.9667802 | 1.300035 | 0.19358892 | 251.4328 | 162.9464 |
| is_heatwave (d=3) | 1.290396 | 0.4966090 | 2.598415 | 0.00936552 | 263.4227 | 64.3140 |
| is_heatwave (d=4) | 1.047806 | 0.7583636 | 1.381667 | 0.16707384 | 185.1389 | 113.4780 |

## A5. Robustness Check - Xgboost & Random forest

### Guangzhou – DML Results for Heatwave Impact on NTL

| Variable | Estimate | Std_Error | t_value | p_value | Percent_Change_NTL | Percent_Change_Se |
|---|---|---|---|---|---|---|
| is_heatwave (d=2) | 0.4841538 | 0.2433256 | 1.989736 | 0.046619988 | 62.28013 | 27.54839 |
| is_heatwave (d=3) | 0.5547622 | 0.2153593 | 2.575984 | 0.009995531 | 74.15267 | 24.03075 |
| is_heatwave (d=4) | 0.3664881 | 0.1681260 | 2.179842 | 0.029269197 | 44.26593 | 18.30857 |

### São Paulo – DML Results for Heatwave Impact on NTL

| Variable | Estimate | Std_Error | t_value | p_value | Percent_Change_NTL | Percent_Change_Se |
|---|---|---|---|---|---|---|
| is_heatwave (d=2) | 0.4150176 | 0.2335926 | 1.776672 | 0.07562219 | 51.43973 | 26.31298 |
| is_heatwave (d=3) | 0.3297625 | 0.1652567 | 1.995456 | 0.04599320 | 39.06378 | 17.96960 |
| is_heatwave (d=4) | 0.3118097 | 0.1274361 | 2.446793 | 0.01441336 | 36.58948 | 13.59123 |

### Delhi - DML Results for Heatwave Impact on NTL

| Variable | Estimate | Std_Error | t_value | p_value | Percent_Change_NTL | Percent_Change_Se |
|----------|----------|-----------|---------|---------|-------------------|-------------------|
| is_heatwave (d=2) | 0.2017630 | 0.3137154 | 0.6431403 | 0.520133058 | 22.35580 | 36.85001 |
| is_heatwave (d=3) | 0.4534919 | 0.1692315 | 2.6797134 | 0.007368522 | 57.37981 | 18.43943 |
| is_heatwave (d=4) | 0.3345436 | 0.2022802 | 1.6538621 | 0.098155543 | 39.73026 | 22.41910 |

### Cairo - DML Results for Heatwave Impact on NTL

| Variable | Estimate | Std_Error | t_value | p_value | Percent_Change_NTL | Percent_Change_Se |
|----------|----------|-----------|---------|---------|-------------------|-------------------|
| is_heatwave (d=2) | 0.5154847 | 0.6074266 | 0.848637 | 0.39608329 | 67.4450 | 83.57014 |
| is_heatwave (d=3) | 0.8454966 | 0.4015492 | 2.105586 | 0.03524029 | 132.9134 | 49.41376 |
| is_heatwave (d=4) | 1.2118980 | 0.5727192 | 2.116042 | 0.03434120 | 235.9856 | 77.30818 |

## A6. Main model summary

ML algorithms: Lasso Regression & Random forest (Regressor)
lasso_learner <- lrn("regr.cv_glmnet")
Rf_learner <- lrn("regr.ranger")

## Guangzhou:

1) d = 2

```
================ DoubleMLPLR Object ================


----------------- Data summary      ------------------
Outcome variable: log_Mean.Area.Weighted.Radiance
Treatment variable(s): is_heatwave
Covariates: tempmax, CDD, humidity, dew, cloudcover, precip, windspeed, solarenergy, CDD_Lag1, CDD_Lag2, heatwave_humidity_interaction,
cloudcover_Lag1, humidity_Lag1, heatwave_flag_solarenergy, tempmax_cloudcover, temp_Lag1, temp_Lag2
Instrument(s):
No. Observations: 2423

----------------- Score & algorithm ------------------
Score function: partialling out
DML algorithm: dml2

----------------- Machine learner   ------------------
ml_l: regr.cv_glmnet
ml_m: regr.ranger

----------------- Resampling        ------------------
No. folds: 10
No. repeated sample splits: 1
Apply cross-fitting: TRUE

----------------- Fit summary       ------------------
 Estimates and significance testing of the effect of target variables
            Estimate. Std. Error t value Pr(>|t|)
is_heatwave   0.5142     0.2730   1.884   0.0596 .
---
Signif. codes:  0 '***' 0.001 '**' 0.01 '*' 0.05 '.' 0.1 ' ' 1
```

## 2) d = 3

```
================ DoubleMLPLR Object ================


----------------- Data summary      ------------------
Outcome variable: log_Mean.Area.Weighted.Radiance
Treatment variable(s): is_heatwave
Covariates: tempmax, CDD, humidity, dew, cloudcover, precip, windspeed, solarenergy, CDD_Lag1, CDD_Lag2, heatwave_humidity_interaction,
cloudcover_Lag1, humidity_Lag1, heatwave_flag_solarenergy, tempmax_cloudcover, temp_Lag1, temp_Lag2
Instrument(s):
No. Observations: 2423

----------------- Score & algorithm ------------------
Score function: partialling out
DML algorithm: dml2

----------------- Machine learner   ------------------
ml_l: regr.cv_glmnet
ml_m: regr.ranger

----------------- Resampling        ------------------
No. folds: 10
No. repeated sample splits: 1
Apply cross-fitting: TRUE

----------------- Fit summary       ------------------
 Estimates and significance testing of the effect of target variables
            Estimate. Std. Error t value Pr(>|t|)
is_heatwave   0.5479     0.2384   2.298   0.0216 *
---
Signif. codes:  0 '***' 0.001 '**' 0.01 '*' 0.05 '.' 0.1 ' ' 1
```

## 3) d = 4

```
================ DoubleMLPLR Object ================


----------------- Data summary      -----------------
Outcome variable: log_Mean.Area.Weighted.Radiance
Treatment variable(s): is_heatwave
Covariates: tempmax, CDD, humidity, dew, cloudcover, precip, windspeed, solarenergy, CDD_Lag1, CDD_Lag2, heatwave_humidity_interaction,
cloudcover_Lag1, humidity_Lag1, heatwave_flag_solarenergy, tempmax_cloudcover, temp_Lag1, temp_Lag2, temp_Lag3, CDD_Lag3
Instrument(s):
No. Observations: 2423

----------------- Score & algorithm -----------------
Score function: partialling out
DML algorithm: dml2

----------------- Machine learner   -----------------
ml_l: regr.cv_glmnet
ml_m: regr.ranger

----------------- Resampling        -----------------
No. folds: 10
No. repeated sample splits: 1
Apply cross-fitting: TRUE

----------------- Fit summary       -----------------
 Estimates and significance testing of the effect of target variables
            Estimate. Std. Error t value Pr(>|t|)
is_heatwave    0.3659     0.1678   2.181   0.0292 *
---
Signif. codes: 0 '***' 0.001 '**' 0.01 '*' 0.05 '.' 0.1 ' ' 1
```

## São Paulo:

1) d = 2

```
================ DoubleMLPLR Object ================


----------------- Data summary      -----------------
Outcome variable: log_Mean.Area.Weighted.Radiance
Treatment variable(s): is_heatwave
Covariates: tempmax, CDD, humidity, dew, cloudcover, precip, windspeed, solarenergy, CDD_Lag1, CDD_Lag2, heatwave_humidity_interaction,
cloudcover_Lag1, humidity_Lag1, heatwave_flag_solarenergy, tempmax_cloudcover, temp_Lag1, temp_Lag2
Instrument(s):
No. Observations: 2433

----------------- Score & algorithm -----------------
Score function: partialling out
DML algorithm: dml2

----------------- Machine learner   -----------------
ml_l: regr.cv_glmnet
ml_m: regr.ranger

----------------- Resampling        -----------------
No. folds: 10
No. repeated sample splits: 1
Apply cross-fitting: TRUE

----------------- Fit summary       -----------------
 Estimates and significance testing of the effect of target variables
            Estimate. Std. Error t value Pr(>|t|)
is_heatwave    0.4765     0.2973   1.603    0.109
```

2) d = 3

```
================= DoubleMLPLR Object =================


----------------- Data summary      -----------------
Outcome variable: log_Mean.Area.Weighted.Radiance
Treatment variable(s): is_heatwave
Covariates: tempmax, CDD, humidity, dew, cloudcover, precip, windspeed, solarenergy, CDD_Lag1, CDD_Lag2, heatwave_humidity_interaction,
cloudcover_Lag1, humidity_Lag1, heatwave_flag_solarenergy, tempmax_cloudcover, temp_Lag1, temp_Lag2
Instrument(s):
No. Observations: 2433

----------------- Score & algorithm -----------------
Score function: partialling out
DML algorithm: dml2

----------------- Machine learner   -----------------
ml_l: regr.cv_glmnet
ml_m: regr.ranger

----------------- Resampling        -----------------
No. folds: 10
No. repeated sample splits: 1
Apply cross-fitting: TRUE

----------------- Fit summary       -----------------
 Estimates and significance testing of the effect of target variables
           Estimate. Std. Error t value Pr(>|t|)
is_heatwave   0.3692     0.1902   1.941   0.0522 .
---
Signif. codes:  0 '***' 0.001 '**' 0.01 '*' 0.05 '.' 0.1 ' ' 1
```

3) d = 4

```
================= DoubleMLPLR Object =================


----------------- Data summary      -----------------
Outcome variable: log_Mean.Area.Weighted.Radiance
Treatment variable(s): is_heatwave
Covariates: tempmax, CDD, humidity, dew, cloudcover, precip, windspeed, solarenergy, CDD_Lag1, CDD_Lag2, heatwave_humidity_interaction,
cloudcover_Lag1, humidity_Lag1, heatwave_flag_solarenergy, tempmax_cloudcover, temp_Lag1, temp_Lag2, temp_Lag3, CDD_Lag3
Instrument(s):
No. Observations: 2433

----------------- Score & algorithm -----------------
Score function: partialling out
DML algorithm: dml2

----------------- Machine learner   -----------------
ml_l: regr.cv_glmnet
ml_m: regr.ranger

----------------- Resampling        -----------------
No. folds: 10
No. repeated sample splits: 1
Apply cross-fitting: TRUE

----------------- Fit summary       -----------------
 Estimates and significance testing of the effect of target variables
           Estimate. Std. Error t value Pr(>|t|)
is_heatwave   0.4665     0.1487   3.138   0.0017 **
---
Signif. codes:  0 '***' 0.001 '**' 0.01 '*' 0.05 '.' 0.1 ' ' 1
```

## Delhi:

1) d = 2

```
================  DoubleMLPLR Object  ================


----------------- Data summary      -----------------
Outcome variable: log_Mean.Area.Weighted.Radiance
Treatment variable(s): is_heatwave
Covariates: tempmax, CDD, humidity, dew, cloudcover, precip, windspeed, solarenergy, CDD_Lag1, CDD_Lag2, heatwave_humidity_interaction,
cloudcover_Lag1, humidity_Lag1, heatwave_flag_solarenergy, tempmax_cloudcover, temp_Lag1, temp_Lag2
Instrument(s):
No. Observations: 2430

----------------- Score & algorithm -----------------
Score function: partialling out
DML algorithm: dml2

----------------- Machine learner   -----------------
ml_l: regr.cv_glmnet
ml_m: regr.ranger

----------------- Resampling        -----------------
No. folds: 10
No. repeated sample splits: 1
Apply cross-fitting: TRUE

----------------- Fit summary       -----------------
 Estimates and significance testing of the effect of target variables
            Estimate. Std. Error t value Pr(>|t|)
is_heatwave    0.1165     0.2807   0.415    0.678
```

2) d = 3

```
================  DoubleMLPLR Object  ================


----------------- Data summary      -----------------
Outcome variable: log_Mean.Area.Weighted.Radiance
Treatment variable(s): is_heatwave
Covariates: tempmax, CDD, humidity, dew, cloudcover, precip, windspeed, solarenergy, CDD_Lag1, CDD_Lag2, heatwave_humidity_interaction,
cloudcover_Lag1, humidity_Lag1, heatwave_flag_solarenergy, tempmax_cloudcover, temp_Lag1, temp_Lag2
Instrument(s):
No. Observations: 2430

----------------- Score & algorithm -----------------
Score function: partialling out
DML algorithm: dml2

----------------- Machine learner   -----------------
ml_l: regr.cv_glmnet
ml_m: regr.ranger

----------------- Resampling        -----------------
No. folds: 10
No. repeated sample splits: 1
Apply cross-fitting: TRUE

----------------- Fit summary       -----------------
 Estimates and significance testing of the effect of target variables
            Estimate. Std. Error t value Pr(>|t|)
is_heatwave   0.13146    0.05644   2.329   0.0199 *
---
Signif. codes:  0 '***' 0.001 '**' 0.01 '*' 0.05 '.' 0.1 ' ' 1
```

3) d = 4

```
================ DoubleMLPLR Object ================


----------------- Data summary      ------------------
Outcome variable: log_Mean.Area.Weighted.Radiance
Treatment variable(s): is_heatwave
Covariates: tempmax, CDD, humidity, dew, cloudcover, precip, windspeed, solarenergy, CDD_Lag1, CDD_Lag2, heatwave_humidity_interaction,
cloudcover_Lag1, humidity_Lag1, heatwave_flag_solarenergy, tempmax_cloudcover, temp_Lag1, temp_Lag2
Instrument(s):
No. Observations: 2430

----------------- Score & algorithm ------------------
Score function: partialling out
DML algorithm: dml2

----------------- Machine learner   ------------------
ml_l: regr.cv_glmnet
ml_m: regr.ranger

----------------- Resampling        ------------------
No. folds: 10
No. repeated sample splits: 1
Apply cross-fitting: TRUE

----------------- Fit summary       ------------------
 Estimates and significance testing of the effect of target variables
            Estimate. Std. Error t value Pr(>|t|)
is_heatwave   0.1972     0.1825   1.081     0.28
```

## Cairo:

1) d = 2

```
================ DoubleMLPLR Object ================


----------------- Data summary      ------------------
Outcome variable: log_Mean.Area.Weighted.Radiance
Treatment variable(s): is_heatwave
Covariates: tempmax, CDD, humidity, dew, cloudcover, precip, windspeed, solarenergy, CDD_Lag1, CDD_Lag2, heatwave_humidity_interaction,
cloudcover_Lag1, humidity_Lag1, heatwave_flag_solarenergy, tempmax_cloudcover, temp_Lag1, temp_Lag2
Instrument(s):
No. Observations: 2436

----------------- Score & algorithm ------------------
Score function: partialling out
DML algorithm: dml2

----------------- Machine learner   ------------------
ml_l: regr.cv_glmnet
ml_m: regr.ranger

----------------- Resampling        ------------------
No. folds: 10
No. repeated sample splits: 1
Apply cross-fitting: TRUE

----------------- Fit summary       ------------------
 Estimates and significance testing of the effect of target variables
            Estimate. Std. Error t value Pr(>|t|)
is_heatwave   1.2568     0.9668    1.3     0.194
```

2) d= 3

```
================ DoubleMLPLR Object ================


----------------- Data summary      -----------------
Outcome variable: log_Mean.Area.Weighted.Radiance
Treatment variable(s): is_heatwave
Covariates: tempmax, CDD, humidity, dew, cloudcover, precip, windspeed, solarenergy, CDD_Lag1, CDD_Lag2, heatwave_humidity_interaction,
cloudcover_Lag1, humidity_Lag1, heatwave_flag_solarenergy, tempmax_cloudcover, temp_Lag1, temp_Lag2
Instrument(s):
No. Observations: 2436

----------------- Score & algorithm -----------------
Score function: partialling out
DML algorithm: dml2

----------------- Machine learner   -----------------
ml_l: regr.cv_glmnet
ml_m: regr.ranger

----------------- Resampling        -----------------
No. folds: 10
No. repeated sample splits: 1
Apply cross-fitting: TRUE

----------------- Fit summary       -----------------
 Estimates and significance testing of the effect of target variables
            Estimate. Std. Error t value Pr(>|t|)
is_heatwave  1.2904     0.4966    2.598   0.00937 **
---
Signif. codes: 0 '***' 0.001 '**' 0.01 '*' 0.05 '.' 0.1 ' ' 1
```

3) d = 4

```
================ DoubleMLPLR Object ================


----------------- Data summary      -----------------
Outcome variable: log_Mean.Area.Weighted.Radiance
Treatment variable(s): is_heatwave
Covariates: tempmax, CDD, humidity, dew, cloudcover, precip, windspeed, solarenergy, CDD_Lag1, CDD_Lag2, heatwave_humidity_interaction,
cloudcover_Lag1, humidity_Lag1, heatwave_flag_solarenergy, tempmax_cloudcover, temp_Lag1, temp_Lag2
Instrument(s):
No. Observations: 2436

----------------- Score & algorithm -----------------
Score function: partialling out
DML algorithm: dml2

----------------- Machine learner   -----------------
ml_l: regr.cv_glmnet
ml_m: regr.ranger

----------------- Resampling        -----------------
No. folds: 10
No. repeated sample splits: 1
Apply cross-fitting: TRUE

----------------- Fit summary       -----------------
 Estimates and significance testing of the effect of target variables
            Estimate. Std. Error t value Pr(>|t|)
is_heatwave  1.0478     0.7584    1.382   0.167
```

## A7. Robustness Check - Combined model summary

Model Summary: Robustness Check -  Xgboost & Random Forest (Classifier)
xgb_learner2 <- lrn("regr.xgboost", nrounds=100, eta=0.1, max_depth=6)
Rf_learner   <- lrn("classif.ranger")

## Guangzhou:

### 1) d=2

```
================= DoubleMLPLR Object =================

------------------ Data summary      ------------------
Outcome variable: log_Mean.Area.Weighted.Radiance
Treatment variable(s): is_heatwave
Covariates: tempmax, CDD, humidity, dew, cloudcover, precip, windspeed, solarenergy, CDD_Lag1, CDD_Lag2, heatwave_humidity_interaction,
cloudcover_Lag1, humidity_Lag1, heatwave_flag_solarenergy, tempmax_cloudcover, temp_Lag1, temp_Lag2
Instrument(s):
No. Observations: 2423

------------------ Score & algorithm ------------------
Score function: partialling out
DML algorithm: dml2

------------------ Machine learner   ------------------
ml_l: regr.xgboost
ml_m: classif.ranger

------------------ Resampling        ------------------
No. folds: 10
No. repeated sample splits: 1
Apply cross-fitting: TRUE

------------------ Fit summary       ------------------
 Estimates and significance testing of the effect of target variables
            Estimate. Std. Error t value Pr(>|t|)
is_heatwave   0.4842     0.2433    1.99   0.0466 *
---
Signif. codes: 0 '***' 0.001 '**' 0.01 '*' 0.05 '.' 0.1 ' ' 1
```

### 2) d = 3

```
================= DoubleMLPLR Object =================

------------------ Data summary      ------------------
Outcome variable: log_Mean.Area.Weighted.Radiance
Treatment variable(s): is_heatwave
Covariates: tempmax, CDD, humidity, dew, cloudcover, precip, windspeed, solarenergy, CDD_Lag1, CDD_Lag2, heatwave_humidity_interaction,
cloudcover_Lag1, humidity_Lag1, heatwave_flag_solarenergy, tempmax_cloudcover, temp_Lag1, temp_Lag2
Instrument(s):
No. Observations: 2423

------------------ Score & algorithm ------------------
Score function: partialling out
DML algorithm: dml2

------------------ Machine learner   ------------------
ml_l: regr.xgboost
ml_m: classif.ranger

------------------ Resampling        ------------------
No. folds: 10
No. repeated sample splits: 1
Apply cross-fitting: TRUE

------------------ Fit summary       ------------------
 Estimates and significance testing of the effect of target variables
            Estimate. Std. Error t value Pr(>|t|)
is_heatwave   0.5548     0.2154    2.576   0.01 **
---
Signif. codes: 0 '***' 0.001 '**' 0.01 '*' 0.05 '.' 0.1 ' ' 1
```

## 3) d = 4

```
================ DoubleMLPLR Object ================


------------------ Data summary      ------------------
Outcome variable: log_Mean.Area.Weighted.Radiance
Treatment variable(s): is_heatwave
Covariates: tempmax, CDD, humidity, dew, cloudcover, precip, windspeed, solarenergy, CDD_Lag1, CDD_Lag2, heatwave_humidity_interaction,
cloudcover_Lag1, humidity_Lag1, heatwave_flag_solarenergy, tempmax_cloudcover, temp_Lag1, temp_Lag2, temp_Lag3, CDD_Lag3
Instrument(s):
No. Observations: 2423

------------------ Score & algorithm ------------------
Score function: partialling out
DML algorithm: dml2

------------------ Machine learner   ------------------
ml_l: regr.xgboost
ml_m: classif.ranger

------------------ Resampling        ------------------
No. folds: 10
No. repeated sample splits: 1
Apply cross-fitting: TRUE

------------------ Fit summary       ------------------
 Estimates and significance testing of the effect of target variables
            Estimate. Std. Error t value Pr(>|t|)
is_heatwave   0.3665      0.1681    2.18   0.0293 *
---
Signif. codes: 0 '***' 0.001 '**' 0.01 '*' 0.05 '.' 0.1 ' ' 1
```

## São Paulo:

## 1) d = 2

```
================ DoubleMLPLR Object ================


------------------ Data summary      ------------------
Outcome variable: log_Mean.Area.Weighted.Radiance
Treatment variable(s): is_heatwave
Covariates: tempmax, CDD, humidity, dew, cloudcover, precip, windspeed, solarenergy, CDD_Lag1, CDD_Lag2, heatwave_humidity_interaction,
cloudcover_Lag1, humidity_Lag1, heatwave_flag_solarenergy, tempmax_cloudcover, temp_Lag1, temp_Lag2
Instrument(s):
No. Observations: 2433

------------------ Score & algorithm ------------------
Score function: partialling out
DML algorithm: dml2

------------------ Machine learner   ------------------
ml_l: regr.xgboost
ml_m: classif.ranger

------------------ Resampling        ------------------
No. folds: 10
No. repeated sample splits: 1
Apply cross-fitting: TRUE

------------------ Fit summary       ------------------
 Estimates and significance testing of the effect of target variables
            Estimate. Std. Error t value Pr(>|t|)
is_heatwave   0.4150      0.2336   1.777   0.0756 .
---
Signif. codes: 0 '***' 0.001 '**' 0.01 '*' 0.05 '.' 0.1 ' ' 1
```

## 2) d = 3

```
================ DoubleMLPLR Object ================


------------------ Data summary      ------------------
Outcome variable: log_Mean.Area.Weighted.Radiance
Treatment variable(s): is_heatwave
Covariates: tempmax, CDD, humidity, dew, cloudcover, precip, windspeed, solarenergy, CDD_Lag1, CDD_Lag2, heatwave_humidity_interaction,
cloudcover_Lag1, humidity_Lag1, heatwave_flag_solarenergy, tempmax_cloudcover, temp_Lag1, temp_Lag2
Instrument(s):
No. Observations: 2433

------------------ Score & algorithm ------------------
Score function: partialling out
DML algorithm: dml2

------------------ Machine learner   ------------------
ml_l: regr.xgboost
ml_m: classif.ranger

------------------ Resampling        ------------------
No. folds: 10
No. repeated sample splits: 1
Apply cross-fitting: TRUE

------------------ Fit summary       ------------------
 Estimates and significance testing of the effect of target variables
           Estimate. Std. Error t value Pr(>|t|)
is_heatwave   0.3298     0.1653   1.995    0.046 *
---
Signif. codes: 0 '***' 0.001 '**' 0.01 '*' 0.05 '.' 0.1 ' ' 1
```

3) d = 4

```
================ DoubleMLPLR Object ================


------------------ Data summary      ------------------
Outcome variable: log_Mean.Area.Weighted.Radiance
Treatment variable(s): is_heatwave
Covariates: tempmax, CDD, humidity, dew, cloudcover, precip, windspeed, solarenergy, CDD_Lag1, CDD_Lag2, heatwave_humidity_interaction,
cloudcover_Lag1, humidity_Lag1, heatwave_flag_solarenergy, tempmax_cloudcover, temp_Lag1, temp_Lag2, temp_Lag3, CDD_Lag3
Instrument(s):
No. Observations: 2433

------------------ Score & algorithm ------------------
Score function: partialling out
DML algorithm: dml2

------------------ Machine learner   ------------------
ml_l: regr.xgboost
ml_m: classif.ranger

------------------ Resampling        ------------------
No. folds: 10
No. repeated sample splits: 1
Apply cross-fitting: TRUE

------------------ Fit summary       ------------------
 Estimates and significance testing of the effect of target variables
           Estimate. Std. Error t value Pr(>|t|)
is_heatwave   0.3118     0.1274   2.447   0.0144 *
---
Signif. codes: 0 '***' 0.001 '**' 0.01 '*' 0.05 '.' 0.1 ' ' 1
```

**Delhi:**

1) d = 2

```
================ DoubleMLPLR Object ================


----------------- Data summary      -----------------
Outcome variable: log_Mean.Area.Weighted.Radiance
Treatment variable(s): is_heatwave
Covariates: tempmax, CDD, humidity, dew, cloudcover, precip, windspeed, solarenergy, CDD_Lag1, CDD_Lag2, heatwave_humidity_interaction,
cloudcover_Lag1, humidity_Lag1, heatwave_flag_solarenergy, tempmax_cloudcover, temp_Lag1, temp_Lag2
Instrument(s):
No. Observations: 2430

----------------- Score & algorithm -----------------
Score function: partialling out
DML algorithm: dml2

----------------- Machine learner   -----------------
ml_l: regr.xgboost
ml_m: classif.ranger

----------------- Resampling        -----------------
No. folds: 10
No. repeated sample splits: 1
Apply cross-fitting: TRUE

----------------- Fit summary       -----------------
 Estimates and significance testing of the effect of target variables
            Estimate. Std. Error t value Pr(>|t|)
is_heatwave   0.2018     0.3137   0.643     0.52
```

2) d= 3

```
================ DoubleMLPLR Object ================


----------------- Data summary      -----------------
Outcome variable: log_Mean.Area.Weighted.Radiance
Treatment variable(s): is_heatwave
Covariates: tempmax, CDD, humidity, dew, cloudcover, precip, windspeed, solarenergy, CDD_Lag1, CDD_Lag2, heatwave_humidity_interaction,
cloudcover_Lag1, humidity_Lag1, heatwave_flag_solarenergy, tempmax_cloudcover, temp_Lag1, temp_Lag2
Instrument(s):
No. Observations: 2430

----------------- Score & algorithm -----------------
Score function: partialling out
DML algorithm: dml2

----------------- Machine learner   -----------------
ml_l: regr.xgboost
ml_m: classif.ranger

----------------- Resampling        -----------------
No. folds: 10
No. repeated sample splits: 1
Apply cross-fitting: TRUE

----------------- Fit summary       -----------------
 Estimates and significance testing of the effect of target variables
            Estimate. Std. Error t value Pr(>|t|)
is_heatwave   0.4535     0.1692   2.68    0.00737 **
---
Signif. codes: 0 '***' 0.001 '**' 0.01 '*' 0.05 '.' 0.1 ' ' 1
```

3) d = 4

```
================ DoubleMLPLR Object ================


----------------- Data summary      -----------------
Outcome variable: log_Mean.Area.Weighted.Radiance
Treatment variable(s): is_heatwave
Covariates: tempmax, CDD, humidity, dew, cloudcover, precip, windspeed, solarenergy, CDD_Lag1, CDD_Lag2, heatwave_humidity_interaction,
cloudcover_Lag1, humidity_Lag1, heatwave_flag_solarenergy, tempmax_cloudcover, temp_Lag1, temp_Lag2
Instrument(s):
No. Observations: 2430

----------------- Score & algorithm -----------------
Score function: partialling out
DML algorithm: dml2

----------------- Machine learner   -----------------
ml_l: regr.xgboost
ml_m: classif.ranger

----------------- Resampling        -----------------
No. folds: 10
No. repeated sample splits: 1
Apply cross-fitting: TRUE

----------------- Fit summary       -----------------
 Estimates and significance testing of the effect of target variables
            Estimate. Std. Error t value Pr(>|t|)
is_heatwave   0.3345     0.2023   1.654   0.0982 .
---
Signif. codes:  0 '***' 0.001 '**' 0.01 '*' 0.05 '.' 0.1 ' ' 1
```

## Cairo:

### 1) d = 2

```
================ DoubleMLPLR Object ================


----------------- Data summary      -----------------
Outcome variable: log_Mean.Area.Weighted.Radiance
Treatment variable(s): is_heatwave
Covariates: tempmax, CDD, humidity, dew, cloudcover, precip, windspeed, solarenergy, CDD_Lag1, CDD_Lag2, heatwave_humidity_interaction,
cloudcover_Lag1, humidity_Lag1, heatwave_flag_solarenergy, tempmax_cloudcover, temp_Lag1, temp_Lag2
Instrument(s):
No. Observations: 2436

----------------- Score & algorithm -----------------
Score function: partialling out
DML algorithm: dml2

----------------- Machine learner   -----------------
ml_l: regr.xgboost
ml_m: classif.ranger

----------------- Resampling        -----------------
No. folds: 10
No. repeated sample splits: 1
Apply cross-fitting: TRUE

----------------- Fit summary       -----------------
 Estimates and significance testing of the effect of target variables
            Estimate. Std. Error t value Pr(>|t|)
is_heatwave   0.5155     0.6074   0.849   0.396
```

### 2) d = 3:

```
================ DoubleMLPLR Object ================


---------------- Data summary      ----------------
Outcome variable: log_Mean.Area.Weighted.Radiance
Treatment variable(s): is_heatwave
Covariates: tempmax, CDD, humidity, dew, cloudcover, precip, windspeed, solarenergy, CDD_Lag1, CDD_Lag2, heatwave_humidity_interaction,
cloudcover_Lag1, humidity_Lag1, heatwave_flag_solarenergy, tempmax_cloudcover, temp_Lag1, temp_Lag2
Instrument(s):
No. Observations: 2436

---------------- Score & algorithm ----------------
Score function: partialling out
DML algorithm: dml2

---------------- Machine learner   ----------------
ml_l: regr.xgboost
ml_m: classif.ranger

---------------- Resampling        ----------------
No. folds: 10
No. repeated sample splits: 1
Apply cross-fitting: TRUE

---------------- Fit summary       ----------------
 Estimates and significance testing of the effect of target variables
            Estimate. Std. Error t value Pr(>|t|)
is_heatwave   0.8455     0.4015   2.106   0.0352 *
---
Signif. codes: 0 '***' 0.001 '**' 0.01 '*' 0.05 '.' 0.1 ' ' 1
```

3) d = 4

```
================ DoubleMLPLR Object ================


---------------- Data summary      ----------------
Outcome variable: log_Mean.Area.Weighted.Radiance
Treatment variable(s): is_heatwave
Covariates: tempmax, CDD, humidity, dew, cloudcover, precip, windspeed, solarenergy, CDD_Lag1, CDD_Lag2, heatwave_humidity_interaction,
cloudcover_Lag1, humidity_Lag1, heatwave_flag_solarenergy, tempmax_cloudcover, temp_Lag1, temp_Lag2
Instrument(s):
No. Observations: 2436

---------------- Score & algorithm ----------------
Score function: partialling out
DML algorithm: dml2

---------------- Machine learner   ----------------
ml_l: regr.xgboost
ml_m: classif.ranger

---------------- Resampling        ----------------
No. folds: 10
No. repeated sample splits: 1
Apply cross-fitting: TRUE

---------------- Fit summary       ----------------
 Estimates and significance testing of the effect of target variables
            Estimate. Std. Error t value Pr(>|t|)
is_heatwave   1.2119     0.5727   2.116   0.0343 *
---
Signif. codes: 0 '***' 0.001 '**' 0.01 '*' 0.05 '.' 0.1 ' ' 1
```



\end{document}